\documentclass[screen, authorversion, acmsmall]{acmart}

\usepackage{amsmath}
\usepackage{graphicx}
\usepackage{hyperref}
\usepackage[utf8]{inputenc}
\usepackage{natbib}
\usepackage{xcolor}
\usepackage{ifthen}
\usepackage[normalem]{ulem}
\usepackage{svg}
\usepackage{subfiles}
\usepackage{xspace}
\usepackage{cleveref}
\usepackage{wrapfig}
\usepackage{dialogue}
\usepackage[most]{tcolorbox}
\usepackage{subcaption}
\usepackage{booktabs}
\usepackage{multirow}
\usepackage{glossaries}
\usepackage{listings}
\usepackage{duckuments}

\usepackage{calc}
\usepackage{tabularray}

\usepackage{calc}

\def\Snospace~{\S{}}

\usepackage{enumitem}

\ifthenelse{\isundefined{\definition}}{\newtheorem{definition}{Definition}}{}
\ifthenelse{\isundefined{\assumption}}{\newtheorem{assumption}{Assumption}}{}
\ifthenelse{\isundefined{\hypothesis}}{\newtheorem{hypothesis}{Hypothesis}}{}
\ifthenelse{\isundefined{\proposition}}{\newtheorem{proposition}{Proposition}}{}
\ifthenelse{\isundefined{\theorem}}{\newtheorem{theorem}{Theorem}}{}
\ifthenelse{\isundefined{\lemma}}{\newtheorem{lemma}{Lemma}}{}
\ifthenelse{\isundefined{\corollary}}{\newtheorem{corollary}{Corollary}}{}
\ifthenelse{\isundefined{\alg}}{\newtheorem{alg}{Algorithm}}{}
\ifthenelse{\isundefined{\example}}{\newtheorem{example}{Example}}{}
\usepackage{lscape}
\usepackage{pifont}% http://ctan.org/pkg/pifont
\newcommand{\cmark}{\ding{51}}%
\newcommand{\xmark}{\ding{55}}%

\newcommand\projectname{{Foundation Model Transparency Index}\xspace}

\newcommand\numindicators{100\xspace}

\newcommand\numdomains{3\xspace}

\newcommand\tldrDone[1]{}

\newcommand\eg{e.g.\xspace}
\newcommand\ie{i.e.\xspace}

\usepackage{authblk}
\makeatletter         
\renewcommand\maketitle{
{\raggedright 
\centering
\vspace{-2.2in}
{\Huge \bfseries \sffamily \@title }%\\[2ex]

\vskip2ex

{\@author}%\\[2ex] 

\vskip2ex

}}
\makeatother

\renewenvironment{abstract}{%
    \itshape
    }%
{}
\begin{document}
\title{Foundation Model Transparency Reports}
\input 
\author[1]{Rishi Bommasani}
\author[1]{Kevin Klyman}
\author[2]{Shayne Longpre}
\author[1]{Betty Xiong}
\author[3]{Sayash Kapoor}
\author[1]{Nestor Maslej}
\author[3]{Arvind Narayanan}
\author[1]{Percy Liang}
\affil[1]{Stanford University}
\affil[2]{Massachusetts Institute of Technology}
\affil[3]{Princeton University\authorcr
\textcolor{white}{}
\authorcr
Stanford Center for Research on Foundation Models (CRFM)
\authorcr
Stanford Institute for Human-Centered Artificial Intelligence (HAI)
}

\renewcommand{\shortauthors}{
Bommasani et al.
}

\maketitle
% Need this hack for compilation to work
\noindent %
% \vspace{-0.2in}
% \medskip
\begin{abstract}
Foundation models are critical digital technologies with sweeping societal impact that necessitates transparency. 
To codify how foundation model developers should provide transparency about the development and deployment of their models, we propose Foundation Model Transparency Reports, drawing upon the transparency reporting practices in social media. 
While external documentation of societal harms prompted social media transparency reports, our objective is to institutionalize transparency reporting for foundation models while the industry is still nascent. 
To design our reports, we identify 6 design principles given the successes and shortcomings of social media transparency reporting.
To further schematize our reports, we draw upon the 100 transparency indicators from the Foundation Model Transparency Index.
Given these indicators, we measure the extent to which they overlap with the transparency requirements included in six prominent government policies (\eg the EU AI Act, the US Executive Order on Safe, Secure, and Trustworthy AI).
Well-designed transparency reports could reduce compliance costs, in part due to overlapping regulatory requirements across different jurisdictions. 
We encourage foundation model developers to regularly publish transparency reports, building upon recommendations from the G7 and the White House.
\end{abstract}

\hypertarget{introduction}{\section{Introduction}}
\label{sec:introduction}

Foundation models are transformative digital technologies \citep{bommasani2021opportunities}, introducing new capabilities \citep{wei2022emergent} and risks \citep{weidinger2022taxonomy} that have garnered unprecedented public attention to AI.
As with earlier digital technologies such as the Internet and social media, the potential for profound societal impact necessitates greater transparency for foundation models.
The 2023 Foundation Model Transparency Index \citep{bommasani2023foundation} confirms that, currently, the foundation model ecosystem is opaque: the Index scored 10 major foundation model developers (\eg OpenAI, Google, Meta) on a 100 point scale for transparency, with developers on average receiving a mere 37 out of 100.

Previous digital technologies, especially social media platforms, have been similarly plagued by insufficient transparency.
Over the past 15 years, social media platforms have come to produce \textit{transparency reports}: public reports, produced on recurring basis, that consolidate information related to usage of their platforms and key platform governance practices like takedown requests and policy enforcement.
Today, transparency reports are an industry standard:
Access Now documents that more than 85 Internet and telecommunications companies have produced transparency reports \citep{accessnow2023transparency}.
The European Union's Digital Services Act mandates transparency reporting for online platforms and formalizes the process to ensure that vital information is publicly reported with sufficient fidelity, frequency, standardization, and accessibility \citep{ec2023consultation}.

While transparency practices are nascent for foundation models, and the current landscape displays both idiosyncratic and systematic opacity \citep{bommasani2023foundation}, governments are stepping in to take corrective measures.
In the United States, Representatives Donald Beyer and Anna Eshoo have introduced the AI Foundation Model Transparency Act to mandate public reporting of standardized information as to be determined by the Federal Trade Commission.
This bill builds on reporting requirements from the October 2023 Executive Order on AI.
In the European Union, the EU AI Act requires transparency on training data, energy usage, model evaluations, and risk management.
Other policies intended to improve transparency include Canada's code of conduct for advanced AI systems, China's generative AI services regulation, and the United Kingdom's request that firms share their responsible scaling policies.

To address the transparency deficits in the foundation model ecosystem, build upon transparency practices for social media platforms, and guide the transparency initiatives proposed by governments, we propose \textit{Foundation Model Transparency Reports}.
Foundation Model Transparency Reports are structured reports that provide essential information about foundation models which developers should publish on a periodic basis.
Such transparency reports would standardize what companies should report, consolidate this information to assist stakeholders in finding it, and structure the information to facilitate subsequent analysis or comparison across multiple developers.
Our transparency reports build upon recommendations under the G7's voluntary code of conduct and the White House's voluntary commitments, both of which state that foundation model developers release transparency reports. 

As we show in \autoref{sec:policy}, while current transparency requirements in government policies often lack precision, our transparency reports specify a precise schema for disclosing information.
In particular, we build on the 100 transparency indicators defined in the Foundation Model Transparency Index \citep{bommasani2023foundation} that concretize transparency for foundation models across the supply chain. 
While \citet{bommasani2023foundation} scored foundation model developers for their existing practices, we describe how developers can implement new reporting practices to inculcate stronger norms of transparency.

Our paper makes three contributions to advance transparency in the foundation model ecosystem.
First, we explore how transparency reporting is conducted in other industries to derive principles; we use these principles to design Foundation Model Transparency Reports.
Second, we align our design with government policies to show how transparency reports could improve compliance and reduce compliance burden across jurisdictions.
Third, we instantiate our design with examples of Foundation Model Transparency Report entries from different foundation models based on publicly available information, setting a clear example for future reports.
Together, our work guides foundation model developers on how to be more transparent and world governments on how to promote transparency through policy.
\hypertarget{background}{\section{Social Media Transparency Reports }}
\label{sec:background}
The rise of social media platforms over the past ten to twenty years provides a natural parallel for foundation models.
Namely, a disruptive and powerful emergent technology came to be widely adopted across societies, thereby intermediating important societal functions such as access to information and interpersonal communication.
Social media has been associated with several types of risk, some of which resulted in substantial societal harms (\eg the Cambridge Analytica scandal, the Rohingya genocide in Myanmar).
These harms are not entirely unrelated to the significant opacity of social media platforms (\eg with respect to how is user data shared, how is contented moderated - necessary but not sufficient), which make it more difficult for governments and external researchers to assess harms to users.
We therefore describe transparency reporting in the context of social media, where it has emerged as a standard practice, to conceptualize transparency reporting for foundation models.

\hypertarget{history}{\subsection{History}}
\label{sec:background-history}
% What is a transparency report
In the context of social media and telecommunications, a \textit{transparency report} is a recurring public report of key metrics  related to legal information and takedown requests, as well as policy and intellectual property enforcement for large online platforms \citep{bankston2017transparency,tspa2023transparency}.
Telecommunications companies and social media platforms have gradually adopted these reports since 2010, in response to public concerns over their handling of privacy, government surveillance, freedom of speech, and misinformation.
Initially, these concerns were triggered by disclosures of dissident information to the Chinese government \citep{schatz2006tech}, the FBI's use of the Patriot Act for surveillance \citep{aclu2010internal}, and Edward Snowden's subsequent disclosures of NSA surveillance practices \citep{greenwald2013nsa}.
% https://www.wsj.com/articles/SB114002162437674809 schatz2006tech
% https://www.aclu.org/press-releases/internal-report-finds-flagrant-national-security-letter-abuse-fbi aclu2010internal
% https://www.theguardian.com/world/2013/jun/06/nsa-phone-records-verizon-court-order greenwald2013nsa
Concerns from users and advertisers would later emerge over moderation practices of harmful content \citep{wfanet2020progress}, spurring greater transparency into platform policy enforcement.
% https://wfanet.org/knowledge/item/2020/09/23/WFA-and-platforms-make-major-progress-to-address-harmful-content wfanet2020progress

% \tldr{Early days of transparency reports.}
To examine how transparency reports emerged, we consider Google's transparency report in 2010.
In 2010, Google first reported government requests for content removal or information \citep{google2010transparency}, as well as where its services were blocked or inaccessible \citep{bankston2017transparency}.
The report showed that Google received over 1200 requests from 36 jurisdictions, providing a greater level of basic insight into the platform (\eg Brazil made 398 requests for over 19000 items to be removed, and Google complied with 68 of these requests). 
% https://googleblog.blogspot.com/2010/04/greater-transparency-around-government.html google2010transparency
Shortly thereafter, LinkedIn, Microsoft, and Twitter began producing their own recurring reports, with an avalanche of adoption following the Snowden revelations in 2013, now including Facebook, Apple, and Yahoo.
Access Now's Transparency Reporting Index documents this increase in adoption: 6 companies produced reports in 2012, compared to over 60 in 2015.
Transparency reporting also gradually expanded in scope to include removals under intellectual property law and the Digital Millennium Copyright Act \citep{accessnow2023transparency}.
Etsy was the first platform, in 2015, to introduce a policy enforcement report, detailing its responses to user violations of its terms of service \citep{etsy2014transparency}.
Since reporting on these incidents could compromise user privacy, companies generally release high-level aggregate statistics. 
% https://extfiles.etsy.com/Press/reports/Etsy_TransparencyReport_2014.pdf etsy2014transparency
Overall, transparency reports came to be an important part of companies' brands and helped foster wider public trust and accountability \citep{bankston2017transparency,tspa2023transparency}. 

% \tldr{Evolution of transparency reports.}
While social media platforms played a significant role in conceptualizing the first versions of transparency reports, civil society organizations drove advances in their scope and utility.
For example, organizations began to rank online platform transparency practices to generate pressure.
The Electronic Frontier Foundation regularly scores corporations on privacy, process, and freedom of speech, in ``Who Has Your Back'' \citep{eff2019whyb}.
% https://www.eff.org/who-has-your-back-2016
In response, high-scoring companies like WordPress and Apple publicized their results \citep{zhu2015perfect}.
% https://wordpress.com/blog/2015/06/17/a-perfect-eff-score-were-proud-to-have-your-back/ zhu2015perfect
Similarly, Ranking Digital Rights maintains its Corporate Accountability Index, to score telecommunications providers on a spectrum of transparency, access and responsibility to users \citep{rdr2019corpacc}.
% https://rankingdigitalrights.org/index2015/
And in response to the Transparency Reporting Toolkit from New America and the Berkman Klein Center \citep{budish2016transparency}, Twitter revamped its reporting standards to follow suggested best practices \citep{kessel2016advancing}.
% https://www.newamerica.org/oti/policy-papers/the-transparency-reporting-toolkit/ budish2016transparency 
% https://blog.twitter.com/official/en_us/a/2016/advancing-transparency-with-more-insightful-data.html kessel2016advancing

% \tldr{Adoption and standardization.}
Subject to the recommendations of civil society, and the associated push for greater accountability, transparency reporting from major social media platforms had evolved to become more interpretable to the general public, more detailed, and more regular in its cadence.
By 2021, 88 technology companies had published transparency reports, with some including downloadable data \citep{accessnow2023transparency}.
For instance, Meta now releases comprehensive and often near-live reports on policy enforcement, intellectual property, government requests, content restrictions, regulatory measures, Internet disruptions, and even widely viewed content \citep{facebook2023facebook}. %so users can see the most widely reached pieces of information
In addition, Meta offers content libraries with APIs for Facebook and Instagram, as well as an ad Library.
However, since 2021 there has been a steep decline in new voluntary transparency reporting from major platforms \citep{rydzak2023stalled};  X, for example, no longer makes updates to its Transparency Center \citep{twitter2023update}.
% https://carnegieendowment.org/2023/11/29/stalled-machines-of-transparency-reporting-pub-91085 rydzak2023stalled
% https://blog.twitter.com/en\_us/topics/company/2023/an-update-on-twitter-transparency-reporting twitter2023update
% Going forward, under the EU's Digital Services Act, large social media platforms are compelled to release certain transparency data at a six month cadence.

\hypertarget{purpose}{\subsection{Purpose}}
\label{sec:background-purpose}
In social media, transparency reporting functions as an instrument for social media companies to make information public.
In particular, these disclosures help alleviate informational deficits on public interest matters spanning privacy, free speech, surveillance, and the reach of harmful content.
Social media platforms are often incentivized to comply with the unethical or secretive requests of governments in order to maintain access to their markets \citep{gorwa_ash_2020}.
While transparency reporting cannot fully deter this incentive, it can inform the public of the scope and extent of a government's intervention into platforms, and spur public pressure as a deterrent to surveillance, censorship, or privacy violations.
% Additionally, transparency reporting around intellectual property challenges and their resolutions can 
Additionally, as social media platforms have been likened to a ``digital public square,'' the processes governing the access and dissemination of speech can have significant societal impact \citep{lazar2023algorithmiccity}.
In light of rising concerns of algorithmic dissemination, echo chambers, and scalable misinformation, transparency reporting could mediate public trust.
In theory, open and transparent processes around speech suppression or amplification would enable a fairer public discourse that is better informed about the measures taken by social media companies to regulate online speech.

Given that the information made available through transparency reports is intrinsically highly multifunctional, transparency reports are simultaneously targeted at a range of stakeholders in the complex platform ecosystem.
Nonexhaustively, these stakeholders include platform users, non-users that are impacted by platform operations, investors, and advertisers.
Users, non-users, and civil society collectively are invested in ensuring that processes that govern platform information are fair and privacy-preserving. 
These concerns are partially addressed by clear documentation of standards and procedures for privacy, compliance with governments, and content moderation, as well as public statistics.
Consequently, civil society organizations have outlined clear criteria by which transparency reports can better satisfy these stakeholder objectives \citep{llanso2021transparency,santaclara2023principles,aspen2021information}.
Similarly, it is in advertisers' interests for their ads to not be associated with offensive or harmful content \citep{wfanet2020progress}.
% https://wfanet.org/knowledge/item/2020/09/23/WFA-and-platforms-make-major-progress-to-address-harmful-content
Certain platforms have regulated political advertising, providing a clear example where monitoring the compliance and impact of company policies can provide a useful basis for academic research by social scientists \citep{edelson2021universal}. 
Lastly, in the absence of corporations supporting public interest research, some have argued that society's abilities to understand and address misinformation, among other harms, is severely limited \citep{abdo2022safe}.

\hypertarget{implementation}{\subsection{Implementation}}
\label{sec:background-implementation}
Modern social media transparency reports are typically divided into four categories: legal information requests, legal takedown requests, intellectual property enforcement, and policy enforcement \citep{tspa2023transparency}. 
Legal information requests typically pertain to requests for private information on users and their communications \citep{Vermeulen2021keys}.
Legal takedown requests pertain to governments applying local laws to have content permanently removed from platforms.
Platforms may not always comply with government requests, so reports often show the number of requests by country, the compliance rate, and the number of unique accounts affected.
Intellectual property reporting is often split into content removals and requests for copyrighted and trademarked content.
Policy enforcement reporting includes a wide range of potential violations, which will differ by platform, and usually display the removal rates over time per country for each violation type.
Additionally, platforms may report other metrics that detail the security of user accounts, the content that is most viewed on the platform, or changes in company policies. 

While this high-level standardization is common across social media companies, further standardization has been challenging.
Primarily, as \citet{keller2021humility} outlines, most metrics are not straightforward to calculate and come with implicit assumptions.
% https://cyberlaw.stanford.edu/blog/2021/03/some-humility-about-transparency keller2021humility
As a result, given the significant heterogeneity in social media platforms, this requires bespoke, company-specific measurement approaches that inhibit apples-to-apples comparisons \citep{tspa2023transparency}.
Because reported statistics are often not standardized, platforms have substantial discretion to select what they measure and how they measure it \citep{UrmanMakhortykh2023}.
This idiosyncratic approach intensifies concerns that transparency reporting in social media serves as a type of ethics-washing that is instrumentalized as marketing collateral \citep{zalnieriute2021transparency}
Further, transparency reporting introduces substantial costs for social media platforms \citep{StoughtonRosenzweig2022}.
Significant company-internal infrastructure is required to initially measure and subsequently maintain metrics, especially given many social media platforms operate across many jurisdictions.
Companies need to dedicate significant resources to maintaining their transparency reports, causing some to question whether this comes at the cost of further investment into more substantive governance or risk mitigation at these organizations.

Serious critiques of transparency reporting---and the broader focus on improving the procedural transparency of digital technology providers---have been raised in various fields \citep{boyd2016algorithmic,ananny2016limits, doi:10.1177/20539517231164119, Mittelstadt2019, han2015transparency,birchall2021radical}. 
For example, social media companies who release transparency reports rarely sufficient access to their platforms for third parties to validate the information they disclose, meaning the information may be inaccurate. 
These critiques are often valid: transparency is not an end unto itself, it is merely a mechanism that may allow further insight into the operations of technology companies in order to better pursue other more tangible societal goals \citep{bommasani2023transparency}. 
The way in which transparency requirements are implemented can have a significant impact on whether transparency is performative and unverifiable or substantive and rigorous.

\hypertarget{mandates}{\subsection{Mandates}}
\label{sec:background-adoption}
Historically, transparency reporting has been a voluntary practice and, increasingly, an expected norm in the social media industry due to public pressure.
However, a growing number of governments are considering mandating transparency reporting.
In the United States, other types of disclosure requirements imposed by the government are at times in tension with the First Amendment due to concerns of compelled speech.
In ruling on disclosure requirements across several contexts, the Supreme Court has used several different legal standards for distinct types of disclosure, sharing a common basis in requiring the government to prove a disclosure requirement ``is appropriately tailored to a sufficiently important goal" \citep{crs2023first}.

Under the European Union's recently-enacted Digital Services Act (DSA), online platforms are required to abide by transparency and access provisions \citep{miller2023tracking}.
The EU designates Very Large Online Platforms (VLOPs) as platforms with at least 45 million monthly active users (\ie 10\% of the EU population): these platforms must prepare biannual transparency reports, conduct periodic risk assessments, publish audit reports, establish ad repositories, and share data with external researchers. 
The EU solicited external input on the form and manner of these transparency reports from December 2023 to January 2024, and intends to adopt an implemented regulation in the first quarter of 2024 \citep{ec2023consultation}.
Primed by the experiences of transparency reporting over the past decade, the EU aims to standardize reporting by identifying a series of indicators that must be reported, along with clarifying measurement methodology in several cases \citep{Schneider2023Collaborative}.
In the first round of transparency reporting under the DSA, 19 platforms submitted reports spanning human resources dedicated to content moderation by locale, content enforcement takedown rates and error rates on both content and accounts, as well as the median time needed to enforce content violating the law or platform policy \citep{dsa2022}.
This level of specificity has allowed for far greater clarity into operations: for example, the transparency report from X demonstrates glaring disparities in content moderation staffing across languages (\eg Bulgarian, Croatian, Dutch, Hebrew, Italian, Latvian and Polish all have at most 2 content moderation staff who are primary language speakers, compared to over 2000 for English).\footnote{See \url{https://transparency.twitter.com/dsa-transparency-report.html}.} 
\hypertarget{design}
{\section{Design of Foundation Model Transparency Reports}}
\label{sec:design}
To design Foundation Model Transparency Reports, we identify 6 design principles, informed directly by the strengths and weaknesses of social media transparency reporting (\autoref{sec:background}).
Subject to these principles, we then identify indicators to be included in the reports using the Foundation Model Transparency Index \citep{bommasani2023foundation} and work through a few examples of how developers may report information related to these indicators. 

\subsection{Principles}
Social media transparency reports, especially in their current form, embody several desirable principles for transparency reporting.
First, these reports consolidate information about a social media platform's practice into a \textit{centralized} location, referring both to the transparency report document and the transparency report page on the platform's website.
Consolidation and centralization enable stakeholders to have a singular and predictable source for finding relevant information.
Second, these reports are \textit{structured} to address specific queries: reports often have four top-level sections (see \autoref{sec:background-implementation}).
This structure sets clear expectations for what can be found in the report, what the report is unlikely to cover, and as a coarse means for comparing different platform practices.
Third, some companies prepare extensive transparency reports that clearly \textit{contextualize} information.
Given that transparency reports are read by a variety of stakeholders with differing expertise and familiarity about platforms, and there are many unique nuances of a platform (\eg what a ``user" is in the context of the platform), context is necessary to adequately interpret information. 

However, social media transparency reports at present (generally) fail to implement other desirable principles for transparency reporting.
First, while these reports consolidate information, the underlying information to be included is not \textit{independently specified}. 
Consequently, platforms are able to determine what information to include and exclude, allowing them to unevenly report only on matters advantageous to them.
Second, while these reports are coarsely structured, they are not fully \textit{standardized} both in terms of the form and organization as well as the indicators reported.
Therefore, transparency reports from different platforms cannot be easily compared to each other or combined to perform larger-scale analyses that reveal aggregate trends.
Social media companies are ultimately the deciders of what constitutes a transparency report in their industry, leading to significant heterogeneity. 
Third, while the best transparency reports at present contextualize information, they often do not clearly specify \textit{methodologies} for computing statistics. 
As a result, given many quantities could be computed in different ways (\eg different methods of user de-duplication for user counting), without clarity on the underlying methodology, consumers of transparency reports may still be prone to misinterpretation.

\subsection{Approach}
Using these 6 principles---centralization, structure, contextualization, independent specification, standardization, and methodologies---we design Foundation Model Transparency Reports.
To begin, rather than having foundation model developers dictate what is included in their own transparency reports, we propose a uniform set of indicators to be included in transparency reports across foundation model developers.
This ensures that the contents of the reports are simultaneously (i) independently specified and (ii) standardized.
To select these indicators, we use the 100 transparency indicators (\autoref{app:indicators}) from the Foundation Model Transparency Index \citep[FMTI;][]{bommasani2023foundation}, which is a recent initiative that scores major foundation model developers for their transparency.
The Foundation Model Transparency Index provides a comprehensive conceptualization of transparency with its 100 indicators organized into 3 domains: (i) the \textit{upstream} resources used to build a foundation model, (ii) the \textit{model} properties including evaluations, and (iii) the \textit{downstream} use and impact of the foundation model.
Domains are further broken down into subdomains (\eg upstream resources include data, labor, compute, code); we re-use this hierarchical domain-subdomain structure as the recommended organization for Foundation Model Transparency Reports.

In contrast to social media platforms, where platform activities and usage are almost exclusively conducted on the platform's website, foundation models have different usage patterns.
As a consequence, transparency reports for foundation models should not only be made available on the foundation model developer's website, but also via distribution channels that make the foundation model available.
For example, Meta's Llama 2 model is distributed via Meta's GitHub repository, but also via Microsoft Azure, Hugging Face, and other platforms.
As a result, transparency reports would, ideally, be disseminated through these distribution channels as well to ensure the associated information can be discovered even if a consumer of the information does not look for it on Meta's website.
Further, akin to the centralization of DSA Transparency Reports in an EU database, governments may consider consolidating Foundation Model Transparency Reports across foundation model developers into a single database to facilitate research and analysis.

Therefore, our design addresses 4 of the 6 principles we identify: independent specification, consolidation/centralization of information, report structure, and standardized information/indicators.
In practice, foundation model developers may choose to not be transparent about certain indicators for a variety of reasons such as (i) the costs of generating the relevant information, (ii) the liability risk from disclosing the information, or (iii) the competitive risk from disclosing the information.
In these cases, we encourage companies to still include these fields in their transparency reports to make clear to other stakeholders this information is not available and, when possible, to justify why this information is not provided.
For example, OpenAI clearly indicates that it is not transparent on several matters (\eg training data, model size) for GPT-4 as a matter of competition and safety \citep{openai2023gpt4}. 

To address the final 2 principles of contextualization and (measurement) methodology, we provide three examples that address indicators across the 3 domains (upstream, model, downstream).

\noindent \textbf{Example: Upstream environmental impact.}
Two of the transparency indicators we include address the direct environmental impact (due to electricity usage) and the broader environmental impact (\eg due to water used to cool data centers) associated with building the foundation model.
As an exemplar of how to provide this transparency, we consider the work of \citet{luccioni2022estimating} in estimating the environmental impact of training BLOOM \citep{scao2022bloom} to underscore three matters.
This work makes clear what is being reported (\ie environmental impacts associated with the equipment manufacturing, model training, and model deployments phases) and what assumptions are made (\eg how different greenhouse gas emissions are converted to tons of carbon dioxide).
Beyond this conceptual clarity, the work provides methodological clarity (\eg in how total emissions are computed as the sum of infrastructure, idle, and dynamic consumption), highlighting components neglected in other environmental accounting approaches \citep{patterson2021carbon}.
Finally, since almost all assessments of environmental impact will hinge on underlying estimates (\eg the carbon intensity of the energy grid), the reporting is clearly contextualized with the sourcing of this information (in this case to statistics provided by Aurora Energy Research on French carbon utilization).

\noindent \textbf{Example: Model evaluations.}
Several of the transparency indicators we include address model evaluations that span capabilities, limitations, risks, mitigations, trustworthiness, and efficiency.
Unlike the environmental impact example, here we instead describe demonstrated issues and challenges in reporting evaluation results with the standard MMLU \citep{hendrycks2021measuring} benchmark for language models.
To ensure evaluation results are correctly interpreted, developers should clearly report the resources involved in adapting (\eg prompting, fine-tuning) their foundation model to the evaluation.
For example, Google reports the results for Gemini \citep{pichai2023gemini} on MMLU in direct comparison to GPT-4, obscuring that Gemini was prompted using 32 examples and chain-of-thought prompting whereas GPT-4 was prompted using 5 examples and standard in-context learning. 
Further, developers should specify lower-level details about model evaluations (\eg the specific prompts used, the codebase and implementation for the evaluation).
\citet{fourrier2023whats} demonstrates that different implementations of MMLU can lead to noticeably different quantitative results, sometimes even changing the ranking of different models.

\noindent \textbf{Example: Downstream policy enforcement.}
Several of the transparency indicators correspond with transparency sought for content moderation on social media platforms. 
Namely, these are indicators on the usage policy for foundation model, the policy's enforcement, the frequency of usage policy violations, the rate of accurate detection of these violations, and whether users are informed about and can appeal moderation decisions.
% Given the direct parallel, social media transparency reporting could provide as a guide.
% In turn, to assess how AI developers should report downstream enforcement, current practices of social media platforms should serve as a bar.
For some of these indicators, producing the relevant information should be of marginal cost to foundation model developers, but we highlight that estimating the rate of usage policy violation is less straightforward. 
Calculating the prevalence of \textit{total} usage policy violations is more involved than just reporting the number of \textit{detected} usage policy violations, since many usage policy violations may go undetected.
To address this issue, \citet{narayanan2023transparencyreports} provide guidance informed by social media practices.
For example, social media companies sample posts uniformly at random to generate estimates of specific policy violations (\eg hate speech) using human moderation.
Foundation model developers could emulate this practice or use other sampling methods to provide better estimates of total violations and detected violations, which would also clarify significant gaps between the the number of total and detected violations.  
\hypertarget{policy-alignment}{\section{Policy Alignment}}
\label{sec:policy}

\begin{table*}[htp]
% \resizebox{\textwidth}{!}{
% \begin{table}
\resizebox{\textwidth}{!}{
  \begin{tabular}{lcccc} 
    \toprule
    Policy & Status & Type & Covered Entities & Reference \\
    \midrule
    {Canada Code of Conduct} & In effect & Voluntary & Firms developing or managing generative AI system with general-purpose capabilities & \citep{canada2023vc} \\
    \addlinespace[1.2ex]  % Tiny space after each row
    EU AI Act & Negotiated & Mandatory & Providers of general purpose AI models, including those with systemic risk & \citep{eu2023aiact} \\
    \addlinespace[1.2ex]
    {G7 Code of Conduct} & In effect & Voluntary & Organizations developing the most advanced AI systems, including the most advanced foundation models & \citep{g72023vc} \\
    \addlinespace[1.2ex]
    US Executive Order & In effect & Mandatory & Firms developing dual-use foundation models trained using >$10^{26}$ FLOPs or $10^{23}$ FLOPs for biological sequence data & \citep{us2023eo} \\
    \addlinespace[1.2ex]
    US FM Transparency Act & Proposed & Mandatory & Providers of foundation models with over 30k monthly users or that generate over 100k monthly output instances & \citep{us2023fmta} \\
    \addlinespace[1.2ex]
    {US White House Commitments} & In effect & Voluntary & Firms developing AI systems & \citep{us2023vc} \\
    \bottomrule
  \end{tabular}}
  \caption{\textbf{Government policies with transparency requirements.} 
Information on the 6 policies we examine: policy name, implementation status as of February 1, 2024, the type of transparency requirements, the entities subject to the requirements, and the reference text we analyze.}
\label{tab:policies}
\end{table*}

\noindent The information that developers can disclose via our transparency reports, in some cases, aligns with requirements by governments. 
We track 6 major policies (\eg the EU AI Act, the US Executive Order on AI), identifying correspondences between our indicators and their requirements. 
Such alignment further incentivizes foundation model developers to report this information (\eg when it is also required by law) and clarifies how different jurisdictions are prioritizing different types of transparency. 
However, the relatively low level of alignment between these policies and our indicators illustrates the lack of granularity in governments' transparency requirements

\subsection{Tracked Policies}
We consider 6 major policies (see \autoref{tab:policies}) from Canada, the EU, the US, and the G7 that include transparency requirements for foundation model developers.
Notably, the US White House voluntary commitments include a pledge that developers will release transparency reports for foundation models that "include the safety evaluations conducted (including in areas such as dangerous capabilities, to the extent that these are responsible to publicly disclose), significant limitations in performance that have implications for the domains of appropriate use, discussion of the model’s effects on societal risks such as fairness and bias, and the results of adversarial testing conducted to evaluate the model’s fitness for deployment."
The G7 Hiroshima Process International Code of Conduct for Organizations Developing Advanced AI Systems includes similar provisions on transparency reports, including transparency regarding evaluations of risks to human rights.

The EU AI Act is a comprehensive regulation covering AI systems that was negotiated in December 2023 and that will be published in spring 2024. 
The AI Act creates a risk taxonomy that imposes requirements on providers of AI systems, prohibits certain use cases, and establishes an AI Office within the European Commission to oversee general-purpose AI systems among many other provisions. Implementation of the AI Act and its transparency requirements will depend significantly on national regulatory authorities within member states, which will be responsible for enforcing the AI Act within domestic legal regimes.  

The US Executive Order on AI, published in October 2023, lays out US policy with respect to attracting AI talent, ensuring security of AI systems, and protecting privacy and civil rights. 
In addition to transparency requirements for some developers of "dual-use foundation models," the order requires that federal government agencies promote competition in the AI industry, establish procedures for the procurement of AI systems, and develop standards for best practices for AI safety. 
The order includes over 150 requirements for federal agencies to complete---nearly all of which must be implemented within one year---meaning that its effects are just beginning to take shape \citep{meinhardt2023tracking}. 

The US AI Foundation Model Transparency Act is a piece of legislation introduced by Representatives Anna Eshoo and Don Beyer in December 2023. The bill is squarely focused on enhancing the transparency of the foundation model ecosystem, ranging from data used for training and inference to transparency standards for foundation model deployers. The US Federal Trade Commission would be tasked with developing and enforcing transparency requirements in consultation with the National Institute for Standards and Technology and the Office of Science and Technology Policy. 

The US White House voluntary commitments are a list of eight commitments made by companies developing AI systems; the White House has announced two rounds of signatories to the commitments, the first by seven companies in July 2023 and the second by eight companies in September 2023.\footnote{The signatories to the US voluntary commitments are Amazon, Anthropic, Google, Inflection, Meta, Microsoft, and OpenAI, which signed in July, and Adobe, Cohere, IBM, Nvidia, Palantir, Salesforce, Scale AI, and Stability AI, which signed in September.} 
In addition to a commitment to publicly release transparency reports for foundation models, the commitments address red teaming, cybersecurity, watermarking, and bias. 
The commitments are not retroactive: they "apply only to generative models that are overall more powerful than the current industry frontier" in the case of companies that signed in July, and "they apply only to generative
models that are overall more powerful than the current most advanced model produced by the company making the commitment" in the case of companies that signed in September.
The US voluntary commitments are intended "to remain in effect until regulations covering substantially the same issues come into force."

The Canada Voluntary Code of Conduct on the Responsible Development and Management of Advanced Generative AI Systems, released in September 2023 by the ministry of Innovation, Science and Economic Development, also introduces a set of nonbinding commitments that has been endorsed by 22 organizations.\footnote{The signatories to the Canada Voluntary Code of Conduct as of February 2024 are Ada, AlayaCare, Alberta Machine Intelligence Institute, AltaML, Appen, BlackBerry, BlueDot, CGI, Cohere, Council of Canadian Innovators, Coveo, IBM, kama.ai, Mila, OpenText, Protexxa Inc., Ranovus, Resemble AI, Responsible Artificial Intelligence Institute, Scale AI, TELUS, and Vector Institute.}
The code of conduct includes specific measures related to different aspects of responsible development and deployment of foundation models, such as accountability, safety, fairness, human oversight, and robustness. 
It directs different measures toward developers and managers of generative AI systems, where managers are organizations that put a system into operation, control access, and conduct monitoring (e.g. developers are responsible for mitigating safety risks, while managers must clearly identify AI-generated content).
Additionally, the code of conduct distinguishes between obligations of developers and managers of all advanced generative AI systems as opposed to those that are made available for public use. 
Similar to the US voluntary commitments, "the code identifies measures that should be applied in advance of binding regulation pursuant to the Artificial Intelligence and Data Act by all firms developing or managing the operations of a generative AI system with general-purpose capabilities."

The G7, which includes the US, Canada, and the EU as members, issued its International Code of Conduct for Organizations Developing Advanced AI Systems in October 2023 as part of the Hiroshima AI Process. 
The code of conduct includes provisions on transparency reporting as well as 10 measures related to data protection, risk management, and development of technical standards. 
While this code of conduct is voluntary and companies have not acceded to it as they have national-level commitments, it may be the basis for a future global agreement.

\subsection{Alignment between existing policies and Foundation Model Transparency Reports}
\begin{table*}[htp]
% \resizebox{\textwidth}{!}{
% \begin{table}
\resizebox{\textwidth}{!}{
\begin{tabular}{lccccc}
\toprule
    Policy & Transparency for whom & \# Upstream & \# Model & \# Downstream & Total \\
    \midrule
    \parbox[t]{5cm}{Canada Code of Conduct} & Public, Firms & 1 & 3 & 5 & 9\\
    \addlinespace[1.2ex]  % Tiny space after each row
    EU AI Act & Public, Firms, Government & 9 & 13 & 8 & 30\\
    \addlinespace[1.2ex]
    \parbox[t]{5cm}{G7 Code of Conduct} & Public & 0 & 7 & 5 & 12\\
    \addlinespace[1.2ex]
    US Executive Order & Government & 0 & 4 & 1 & 5\\
    \addlinespace[1.2ex]
    US FM Transparency Act & Public & 10 & 7 & 3 & 20\\
    \addlinespace[1.2ex]
    \parbox[t]{5cm}{US White House Commitments} & Public & 0 & 6 & 1 & 7\\
    \bottomrule
  \end{tabular}}
  \caption{\textbf{Alignment between government policies and our Foundation Model Transparency Reports.}
For each policy, we indicate which entities receive the disclosed information as well as the overlap between the policy's requirements and our Transparency Report indicators.
We report the overlap in aggregate as well as for (i) upstream resources, (ii) model-level properties, and (iii) downstream use.
Overall, the transparency requirements in all 6 policies are considerably less comprehensive and less specific than the 100 indicators we consider.
}
\label{tab:policy-alignment}
% \end{table}
\end{table*}

\noindent 
To measure the alignment between our transparency reports and existing policy, we tag the transparency requirements in each policy and identify alignment with specific indicators in for every transparency requirement that corresponds with our transparency indicators (see \autoref{app:alignment} for further details).
On average, the 6 policies share 10 transparency requirements with our 100 transparency indicators, and across all policies there are 43 transparency requirements shared with our transparency indicators (see \autoref{tab:policy-alignment}).
That is, there are 57 transparency indicators we include that are not included in any of these policies.
Relatively few of these transparency requirements focus on the upstream resources required to build foundation models, such as data, labor, and compute. 
3 of the 6 policies include no upstream requirements, though the US AI Foundation Model Transparency Act has 10 such requirements, including disclosure of data size, data sources, data augmentation, and personal information in the data.
The EU AI Act had the most transparency requirements, including 12 requirements that no other policy contained such as the duration of model development, energy usage, model components, and the model license.  
The US Executive Order on AI had the fewest transparency requirements of all policies considered and, notably, was the sole policy to require only that companies disclose information to the government.

Common transparency requirements across policies included disclosing data sources (required by 3 policies), centralized model documentation (3), prohibited, restricted, and prohibited uses (3), whether a person is interacting with an AI system (3), documentation for responsible downstream use (3), a capabilities description (4), a risk description (4), evaluation of unintentional harm (4), evaluation of intentional harm (5), a limitations description (5), and a mitigations description (5).
These commonalities show shared priorities, but also that current transparency requirements are often superficial; in particular, our indicators include not only descriptions of capabilities, limitations, risks, and mitigations, but also demonstrations and evaluations of each.
In the upstream domain, the only consistent transparency requirement relates to data sources, with no policies including transparency requirements related to data labor and only one policy (the AI Act) with a requirement on technical methods.

There were also a handful of transparency requirements that are included in these policies that were not featured in the Foundation Model Transparency Index. 
For example, Canada's code of conduct requires that developers "maintain a database of reported incidents after deployment, and provide updates as needed to ensure effective mitigation measures." 
It also requires that firms that manage the operations of generative AI systems "share information and best practices on risk management with firms playing complementary roles in the ecosystem." 
The EU AI Act contains a number of additional transparency requirements, ranging from the date the model was released to the maximum context window length, the rationale for key design choices, and adverse event reporting. 
We revisit adverse event reporting, which appears in both of these policies, in \autoref{sec:other-reporting}

On the whole, transparency requirements in government policies lack specificity; they do not detail the precision to which developers must report quantitative information, establish standards for reporting evaluations, or account for differences across modalities.
Foundation Model Transparency Reports may help augment vague government policies by sharpening what information foundation model developers provide to consumers, clients, downstream developers, deployers, and regulators.
\hypertarget{example}
{\section{Example of Transparency Report Entries}}
\label{sec:example}

To demonstrate how to construct a Foundation Model Transparency Report, we provide an example of transparency report entries in \autoref{app:example}.
For brevity, here we describe how we assembled these examples and takeaways given current disclosure practices. 

\subsection{Construction}
The 2023 Foundation Model Transparency Index (FMTI) confirms that current transparency practices across the foundation model ecosystem are lackluster.
Most major foundation model developers do not provide information on over half of our 100 indicators \citep{bommasani2023foundation}.
As a result, for the purposes of building an illustrative sample report, we provide examples of transparency report entries from 9 foundation model developers instead of reporting on the practices of a single developer.
While in practice, a transparency report will correspond to the practices of a single developer (and be associated with a single model or model family), we nonetheless believe this amalgam serves a useful demonstration.

To create these examples, we consider the 10 foundation model developers scored in FMTI and the associated scores.
For any indicator (82 of the 100) where at least one developer scores the point, we consider all developers that receive a point for that indicator.
Of these developers, we select one of the developers whose practices best exemplify transparency for the indicator, with some consideration for selecting different developers to portray a variety of practices.
Given the selected (indicator, developer) pairs, we then prepared the example entry in \autoref{app:example} that articulates what the developer discloses for the indicator (\eg Meta's disclosure of development duration for Llama 2, Inflection's description of the limitations of Inflection-1).
While in our judgment this report reflects some of the best existing practices for each indicator, we note that this should not be seen as the ceiling for transparency in many cases.

\subsection{Analysis}
Our examples of Foundation Model Transparency Report entries, in line with findings of the FMTI, implicitly denotes 18 indicators where no major developer is currently transparent (\eg several labor-related indicators, several indicators on usage statistics and impact).
Consequently, developers or others in the community that demonstrate how information regarding these 18 indicators should be disclosed would establish a meaningful precedent. 
Further, even for many of the 82 indicators where the report contains an entry, significantly more could be done to make this information useful and actionable.
In many cases, the level of contextualization and methodological clarity could be specifically improved (\eg regarding the aspects of model development that contributed to Meta's measurement of duration, or specific limitations were identified by Inflection?).

At a more fine-grained level, certain indicators also reveal how foundation model developers conceptualize model development differently, and the community lacks a common conceptual framework.
For example, while practices from Anthropic, Hugging Face/BigScience, Meta, and OpenAI are all included in the example report on the matters of data and labor, different developers describe their data pipelines in substantively different ways.
In turn, articulating where human labor is involved and what data processing occurs through this pipeline may yield inconsistent answers across developers that may not be directly comparable.
For many indicators in the report, it is unclear if the disclosed information is a partial or complete answer.
For example, while policies from Anthropic, Google, Inflection and OpenAI are all included on the matters of terms of service and usage policy enforcement, in several cases it remains unclear whether they capture an exhaustive list of violative behavior, enforcement actions, and associated appeals/justifications. 
In fact, in some cases this information was only identified by triggering detected usage policy violations by \citet{bommasani2023foundation}, which brings into question the extent to which these usage policies are fully transparent at present.
\hypertarget{related}
{\section{Related Work}}
\label{sec:related}
Transparency is a fundamental value with a significant history of study in AI \citep{gebru2021datasheets, bender2018data, mitchell2018modelcards, raji2019actionable, gray2019ghost, crawford2021atlas, cdt2021, keller2022platform, bommasani2023ecosystem}.
Here we consider how our approach relates to other transparency methodologies in AI (namely model cards, data sheets, and ecosystem cards) and other reporting methodologies in society (namely financial reporting and adverse event reporting). 
While Foundation Model Transparency Reports draw greatest inspiration from social media transparency reports (\autoref{sec:background}), these other methodological approaches to transparency and reporting can help inform and complement more comprehensive transparency reporting.

\subsection{Transparency approaches in AI}
The most common approach for improving in transparency in AI is model \textit{evaluations}: these evaluations help to clarify model strengths and weaknesses, often for technical AI practitioners \citep{bommasani2023transparency}.
While evaluations can provide significant insight into a specific model, they are still limited in their ability to account for broader societal context (\eg data, labor, downstream impact).
In turn, \textit{documentation}-based approaches to increasing transparency play a complementary role to evaluations, often providing legibility to stakeholders beyond technical AI practitioners.
While model evaluations characterize a specific model in isolation, documentation situates model and system development in a broader context.

Documentation in AI was pioneered by data sheets \citep{gebru2018datasheets} and model cards \citep{mitchell2018modelcards} for data and models, respectively.
These documentation frameworks enumerate a series of questions that an AI developer should answer, which are often fairly open-ended and unstructured in form.
For example, \citet{gebru2018datasheets} introduces three questions on the motivation for dataset creation: what was the purpose for dataset creation, who created the dataset (and, potentially on whose behalf), and who funded the dataset?
These documentation approaches tend to be very comprehensive in their maximal instantiation, which means empirically there is significant heterogeneity in how different organizations produce data sheets or model cards, including which of the original questions posed by \citet{gebru2018datasheets} and \citet{mitchell2018modelcards} are (satisfactorily) addressed. 
For example, the Llama 2 model card contains most of the high-level categories specified in the original model cards paper, but several of the lower-level questions posed in the paper are not addressed. Another important example of documentation in AI are the reproducibility checklists required by conferences like NeurIPS and EMNLP, which are mandatory for all papers and include various transparency requirements related to training, licensing, and limitations \citep{NeurIPS2022, EMNLP2023}.

More recently, \citet{bommasani2023ecosystem} introduced ecosystem cards as a documentation framework, which akin to our work specifically targets the foundation model setting.
Three variants of the ecosystem card template exist for documenting datasets, foundation models, and applications/products respectively, with \citet{bommasani2023ecosystem} emphasizing the importance of tracking dependency relationships between these different assets. 
In contrast to data sheets and model cards, which were principally envisioned as developer-driven forms of transparency, ecosystem cards can be created and maintained by other actors in the ecosystem.

Relative to these documentation frameworks, Foundation Model Transparency Reports share common themes of organizing information and, in several instances, specific indicators.
However, our transparency reports adopt the more comprehensive view of transparency put forth in the Foundation Model Transparency Index, spanning elements across the supply chain.
Further, our transparency reports are closer in style to social media transparency reports, with a greater emphasis on more targeted informational queries rather than more open-ended questions found in data sheets and model cards.
Our focus is on transparency that is relevant for public accountability and risk management in relation to (widely-deployed) foundation models, whereas many of these prior frameworks are aimed at AI researchers to promote better scientific practices.

\subsection{Reporting approaches in society}
\label{sec:other-reporting}
In mature industries, companies and organizations are often required to produce reports that document their operations (\eg tax reporting, environmental reporting, product safety reporting).
We consider US financial reporting as a horizontal practice spanning industries, as well as the US Food and Drug Administration's (FDA) adverse event reporting system as a domain-specific practice.
These reporting approaches, along with social media transparency reporting, provide additional references in envisioning, designing, and implementing transparency reporting for foundation models. \\
\noindent \textbf{Financial reporting.}
In the United States, several overlapping reporting mechanisms provide transparency on the financial ecosystems.
Financial reporting is overseen by the Securities and Exchange Commission (SEC), whose mandate is to inform and protect investors, regulate securities markets, and enforce federal securities law \citep{SECmain}. 
The SEC requires that publicly traded companies release significant information about their finances through annual reports (Form 10-K), quarterly reports (Form 10-Q), and current reports (Form 8-K) \citep{SECform8k}. 
The 10-K and 10-Q comprehensively characterize a company's financial health (\eg information on business activities, risk factors, assets, liabilities) \citep{SECform10K}, whereas the 8-K is required to notify the SEC, and later the public, of sudden events such as bankruptcy	or acquisition of significant assets \citep{SECform8k}.
% . The Form 10-Q, likewise comprehensive, is a quarterly report of a company’s financial performance \citep{SECform10Q}. However unlike the Form 10-K, the Form 10-Q is unaudited \citep{SECform10Q}. Finally the Form 8-K is a form that public companies must submit to notify the SEC, and consequently the publicm about important sudden events, such as a bankruptcy	or acquisition of significant assets \citep{SECform8k}.
Standards also heavily influence financial reporting.
The Generally Accepted Accounting Principles determine accounting standards accepted by the SEC and function as the default for American companies \citep{GAAP}, with the International Financial Reporting Standards functioning as their international counterpart.
% \citep{ortega2017converg, harris2022converg}.
Additionally, the non-profit Public Company Accounting Oversight Board (PCAOB) develops auditing standards for public companies and SEC-registered brokers and dealers. 
These standards are important as they ensure that business audits are standardized, high quality, and trustworthy \citep{PAOCB}.

The history of US financial regulation has several instructive lessons for transparency reporting for foundation models. 
Many American financial reporting mechanisms came out of regulatory measures intended to address issues of low transparency and their subsequent negative effects. 
The SEC was created in 1934 by Franklin Delano Roosevelt in the aftermath of the 1932 Pechora Commission, which highlighted how abusive practices in the financial industry contributed to the 1929 stock market crash \citep{perino2010hellhound}. 
Likewise, the PCAOB was created as part of the 2002 Sarbanes-Oxley Act, which was itself a response to major corporate accounting scandals like Enron, WorldCom and Tyco.
Beyond creating the PCAOB, the Sarbanes-Oxley Act instrumentally reformed corporate governance and financial disclosure practices in the US, mandating greater financial disclosure, stricter internal corporate control, and greater corporate responsibility over financial reporting.
% \citep{SarbanesOxleyLegacy2023}. 
Therefore there is a well-established precedent of government intervention as a means of ensuring greater transparency in industries that are deemed to be insufficiently transparent.  \\
\noindent \textbf{FDA adverse event reporting.}
While transparency often aims to provide baseline understanding and information, sometimes further transparency is necessary in light of unexpected circumstances.
In the context of drugs, the FDA implements an adverse event reporting system (FAERS) as a ``database that contains information on adverse event and medication error reports submitted to FDA. 
The database is designed to support the FDA's post-marketing safety surveillance program for drug and therapeutic biologic products" \citep{dashboard, faers}.
As of September 2023, there are more than 27 million reports, with the FDA receiving more than one million reports annually since 1969 \citep{dashboard}; the FAERS data is made available to wide range of stakeholders (\eg consumers, healthcare professionals, researchers). 
% With quarterly updates of data files, FAERS has also expanded into a Public Dashboard, FAERS, such a tool expands access of FAERS data to the general public (stakeholders include consumers, healthcare professionals, researchers) to look for reporting related to adverse events \citep{dashboard}. Apart from the dashboard, the general public can obtain FAERS data through FAERS data files and individual case safety reports \citep{faersqa}.

Reports are voluntarily submitted by healthcare providers (\eg physicians, pharmacists, nurses) and consumers (\eg patients, family members, lawyers); by law, product manufacturers must relay these reports to the FDA \citep{faersqa}.
Reports are circulated and may trigger subsequent actions (\eg evaluation by clinical reviewers in the Center for Drug Evaluation and Research) to survey post-market drug safety.
Overall, the open availability of FAERS data improves awareness of drug adverse events, though it may be prone to improper interpretation without appropriate consideration for statistical validity \citep{faersimplication}.
In comparison to the recurring, comprehensive, and proactive nature of social media transparency reports or financial reports, adverse event reporting systems provide more targeted transparency when interventions are (potentially) urgent.
While our focus in designing transparency reports for foundation models largely emulates the former approaches, we highlight adverse event reporting as playing a potentially complementary role.
In particular, we imagine that as specific harms of foundation models are documented, similar adverse event reporting systems (or the reuse of pre-existing systems) will be necessary \citep{guha2023ai, naiac2023aers}.

\hypertarget{discussion}{\section{Discussion}}
\label{sec:discussion}
Transparency functions as an instrument for advancing other objectives (\eg greater public accountability and improved risk management).
We aim to inculcate robust norms and industry standards around transparency while foundation models are still (relatively) nascent, in conjunction with government-driven disclosure requirements.
Transparency is not a monolith: different aspects of transparency are more relevant for certain societal objectives and stakeholder groups than others.
While in some cases the benefits of transparency arise from a single developer being more transparent, for others what is required is broader transparency from many developers to surface general trends.
We step through several of our transparency indicators to articulate our theory of change regarding how increased transparency would help improve the societal impact of foundation models.

Greater transparency on data directly informs demographic biases in foundation model behavior \citep{abid2021persistent, luccioni2023stable, bianchi2023easily} and copyright litigation surrounding model training data \citep[\eg][]{nyt2023nyt}.
Transparency on labor practices enables awareness of, and collective action to address, labor conditions \citep{williams2022exploited}.
Transparency on compute usage clarifies the costs of building frontier foundation models \citep{anderljung2023frontier} and the viability of policies like licensing that restrict compute access \citep{kapoor2023licensing}.
Evaluations help concretize model capabilities \citep{wei2022emergent, bubeck2023sparks} and risks \citep{bender2021dangers, weidinger2022taxonomy}, sharpening collective understanding \citep{bommasani2023transparency}.
And transparency on usage statistics as well as affected market sectors and geographies directly informs understanding of economic impact, innovation, and the concentration of power \citep{vipra2023concentration, bommasani2023ecosystem, cma2023ai}.

While we advocate for greater transparency via transparency reports, we recognize that transparency initiatives have been subject to critique \citep{ananny2016limits, bates2023socially}.
Though some of these critiques regarding performative transparency on self-selected matters are mitigated by our approach \citep{zalnieriute2021transparency}, other critiques about the limits of transparency to bring about substantive change persist \citep{hartzog2023oversight}.
We see improved transparency as a natural initial target given the demonstrated opacity of the foundation model ecosystem \citep{perrigo2022kenya, hao2023cleaning, bommasani2023foundation}; other changes will need to follow to achieve better societal outcomes.
Further, social media transparency reporting demonstrates that transparency reporting can be costly, requiring substantial investments from platforms.
At present, we do not aim to factor in reporting costs, though we encourage developers to transparently discuss costs to allow policymakers and other stakeholders to better argue for cost-benefit trade-offs for transparency.
For similar reasons, we also highlight the potential for Foundation Model Transparency Reports to reduce overall compliance costs for developers operating across multiple jurisdictions by reducing duplicative effort (\autoref{sec:policy}). 

Broad consensus exists for improved transparency in the foundation model ecosystem.
The history of social media illustrates both the harms of pervasive opacity and the potential for institutionalized transparency.
We envisage Foundation Model Transparency Reports as the structured interface for communicating information from foundation model developers to the public to meet the needs of diverse stakeholders.
% \hypertarget{conclusion}%{\section{Conclusion}}
%\label{sec:conclusion}
%Broad consensus exists for improved transparency in the foundation model ecosystem.
%The history of social media illustrates both the harms of pervasive opacity and the potential for institutionalized transparency.
%We envisage Foundation Model Transparency Reports as the structured interface for communicating information from foundation model developers to the public to meet the needs of diverse stakeholders.
%\pl{I don't think we need a separate section for this conclusion - you could just include it as the last paragraph in the discussion}
\begin{acks}
We thank Dan Ho, Daphne Keller, and Nate Persily for feedback and discussions that informed this work.
This work was supported in part by the AI2050 program at Schmidt Futures (Grant G-22-63429).
\end{acks}
 
\clearpage
\bibliographystyle{ACM-Reference-Format}
\bibliography{refdb/all, main}

%%% -*-BibTeX-*-
%%% Do NOT edit. File created by BibTeX with style
%%% ACM-Reference-Format-Journals [18-Jan-2012].

\begin{thebibliography}{99}

%%% ====================================================================
%%% NOTE TO THE USER: you can override these defaults by providing
%%% customized versions of any of these macros before the \bibliography
%%% command.  Each of them MUST provide its own final punctuation,
%%% except for \shownote{}, \showDOI{}, and \showURL{}.  The latter two
%%% do not use final punctuation, in order to avoid confusing it with
%%% the Web address.
%%%
%%% To suppress output of a particular field, define its macro to expand
%%% to an empty string, or better, \unskip, like this:
%%%
%%% \newcommand{\showDOI}[1]{\unskip}   % LaTeX syntax
%%%
%%% \def \showDOI #1{\unskip}           % plain TeX syntax
%%%
%%% ====================================================================

\ifx \showCODEN    \undefined \def \showCODEN     #1{\unskip}     \fi
\ifx \showDOI      \undefined \def \showDOI       #1{#1}\fi
\ifx \showISBNx    \undefined \def \showISBNx     #1{\unskip}     \fi
\ifx \showISBNxiii \undefined \def \showISBNxiii  #1{\unskip}     \fi
\ifx \showISSN     \undefined \def \showISSN      #1{\unskip}     \fi
\ifx \showLCCN     \undefined \def \showLCCN      #1{\unskip}     \fi
\ifx \shownote     \undefined \def \shownote      #1{#1}          \fi
\ifx \showarticletitle \undefined \def \showarticletitle #1{#1}   \fi
\ifx \showURL      \undefined \def \showURL       {\relax}        \fi
% The following commands are used for tagged output and should be
% invisible to TeX
\providecommand\bibfield[2]{#2}
\providecommand\bibinfo[2]{#2}
\providecommand\natexlab[1]{#1}
\providecommand\showeprint[2][]{arXiv:#2}

\bibitem[goo(2010)]%
        {google2010transparency}
 \bibinfo{year}{2010}\natexlab{}.
\newblock \bibinfo{title}{Greater transparency around government requests}.
\newblock \bibinfo{howpublished}{\url{https://googleblog.blogspot.com/2010/04/greater-transparency-around-government.html}}.
\newblock


\bibitem[acl(2010)]%
        {aclu2010internal}
 \bibinfo{year}{2010}\natexlab{}.
\newblock \bibinfo{title}{Internal Report Finds Flagrant National Security Letter Abuse By FBI}.
\newblock \bibinfo{howpublished}{\url{https://www.aclu.org/press-releases/internal-report-finds-flagrant-national-security-letter-abuse-fbi}}.
\newblock


\bibitem[ets(2014)]%
        {etsy2014transparency}
 \bibinfo{year}{2014}\natexlab{}.
\newblock \bibinfo{title}{2014 Transparency Report}.
\newblock \bibinfo{howpublished}{\url{https://extfiles.etsy.com/Press/reports/Etsy_TransparencyReport_2014.pdf}}.
\newblock


\bibitem[wfa(2020)]%
        {wfanet2020progress}
 \bibinfo{year}{2020}\natexlab{}.
\newblock \bibinfo{title}{WFA and platforms make major progress to address harmful content}.
\newblock \bibinfo{howpublished}{\url{https://wfanet.org/knowledge/item/2020/09/23/WFA-and-platforms-make-major-progress-to-address-harmful-content}}.
\newblock


\bibitem[Abdo et~al\mbox{.}(2022)]%
        {abdo2022safe}
\bibfield{author}{\bibinfo{person}{Alex Abdo}, \bibinfo{person}{Ramya Krishnan}, \bibinfo{person}{Stephanie Krent}, \bibinfo{person}{Evan Welber~Falc{\'o}n}, {and} \bibinfo{person}{Andrew~Keane Woods}.} \bibinfo{year}{2022}\natexlab{}.
\newblock \bibinfo{title}{A Safe Harbor for Platform Research}.
\newblock \bibinfo{howpublished}{\url{https://knightcolumbia.org/content/a-safe-harbor-for-platform-research}}.
\newblock


\bibitem[Abid et~al\mbox{.}(2021)]%
        {abid2021persistent}
\bibfield{author}{\bibinfo{person}{Abubakar Abid}, \bibinfo{person}{Maheen Farooqi}, {and} \bibinfo{person}{James Zou}.} \bibinfo{year}{2021}\natexlab{}.
\newblock \showarticletitle{Persistent anti-muslim bias in large language models}.
\newblock \bibinfo{journal}{\emph{arXiv preprint arXiv:2101.05783}} (\bibinfo{year}{2021}).
\newblock


\bibitem[{Access Now}(2023)]%
        {accessnow2023transparency}
\bibfield{author}{\bibinfo{person}{{Access Now}}.} \bibinfo{year}{2023}\natexlab{}.
\newblock \bibinfo{title}{Transparency Reporting Index}.
\newblock \bibinfo{howpublished}{\url{https://www.accessnow.org/campaign/transparency-reporting-index/}}.
\newblock


\bibitem[Ananny and Crawford(2018)]%
        {ananny2016limits}
\bibfield{author}{\bibinfo{person}{Mike Ananny} {and} \bibinfo{person}{Kate Crawford}.} \bibinfo{year}{2018}\natexlab{}.
\newblock \showarticletitle{Seeing without knowing: Limitations of the transparency ideal and its application to algorithmic accountability}.
\newblock \bibinfo{journal}{\emph{New Media \& Society}} \bibinfo{volume}{20}, \bibinfo{number}{3} (\bibinfo{year}{2018}), \bibinfo{pages}{973--989}.
\newblock
\urldef\tempurl%
\url{https://doi.org/10.1177/1461444816676645}
\showDOI{\tempurl}
\showeprint{https://doi.org/10.1177/1461444816676645}


\bibitem[Anderljung et~al\mbox{.}(2023)]%
        {anderljung2023frontier}
\bibfield{author}{\bibinfo{person}{Markus Anderljung}, \bibinfo{person}{Joslyn Barnhart}, \bibinfo{person}{Anton Korinek}, \bibinfo{person}{Jade Leung}, \bibinfo{person}{Cullen O'Keefe}, \bibinfo{person}{Jess Whittlestone}, \bibinfo{person}{Shahar Avin}, \bibinfo{person}{Miles Brundage}, \bibinfo{person}{Justin Bullock}, \bibinfo{person}{Duncan Cass-Beggs}, \bibinfo{person}{Ben Chang}, \bibinfo{person}{Tantum Collins}, \bibinfo{person}{Tim Fist}, \bibinfo{person}{Gillian Hadfield}, \bibinfo{person}{Alan Hayes}, \bibinfo{person}{Lewis Ho}, \bibinfo{person}{Sara Hooker}, \bibinfo{person}{Eric Horvitz}, \bibinfo{person}{Noam Kolt}, \bibinfo{person}{Jonas Schuett}, \bibinfo{person}{Yonadav Shavit}, \bibinfo{person}{Divya Siddarth}, \bibinfo{person}{Robert Trager}, {and} \bibinfo{person}{Kevin Wolf}.} \bibinfo{year}{2023}\natexlab{}.
\newblock \bibinfo{title}{Frontier AI Regulation: Managing Emerging Risks to Public Safety}.
\newblock
\newblock
\showeprint[arxiv]{2307.03718}~[cs.CY]


\bibitem[{Aspen Institute}(2021)]%
        {aspen2021information}
\bibfield{author}{\bibinfo{person}{{Aspen Institute}}.} \bibinfo{year}{2021}\natexlab{}.
\newblock \bibinfo{title}{Commission on Information Disorder Final Report}.
\newblock \bibinfo{howpublished}{\url{https://www.aspeninstitute.org/wp-content/uploads/2021/11/Aspen-Institute_Commission-on-Information-Disorder_Final-Report.pdf}}.
\newblock


\bibitem[Bankston et~al\mbox{.}(2017)]%
        {bankston2017transparency}
\bibfield{author}{\bibinfo{person}{Kevin Bankston}, \bibinfo{person}{Ross Schulman}, {and} \bibinfo{person}{Liz Woolery}.} \bibinfo{year}{2017}\natexlab{}.
\newblock \bibinfo{title}{Case Study \#3: Transparency Reporting}.
\newblock \bibinfo{howpublished}{\url{https://www.newamerica.org/in-depth/getting-internet-companies-do-right-thing/case-study-3-transparency-reporting/}}.
\newblock


\bibitem[Bates et~al\mbox{.}(2023)]%
        {bates2023socially}
\bibfield{author}{\bibinfo{person}{J. Bates}, \bibinfo{person}{H. Kennedy}, {and} \bibinfo{person}{I.~et~al. Medina~Perea}.} \bibinfo{year}{2023}\natexlab{}.
\newblock \showarticletitle{Socially meaningful transparency in data-based systems: reflections and proposals from practice}.
\newblock \bibinfo{journal}{\emph{Journal of Documentation}} (\bibinfo{year}{2023}).
\newblock
\showISSN{0022-0418}
\urldef\tempurl%
\url{https://doi.org/10.1108/JD-01-2023-0006}
\showDOI{\tempurl}


\bibitem[Bender and Friedman(2018)]%
        {bender2018data}
\bibfield{author}{\bibinfo{person}{Emily~M Bender} {and} \bibinfo{person}{Batya Friedman}.} \bibinfo{year}{2018}\natexlab{}.
\newblock \showarticletitle{Data statements for natural language processing: Toward mitigating system bias and enabling better science}.
\newblock \bibinfo{journal}{\emph{Transactions of the Association for Computational Linguistics (TACL)}}  \bibinfo{volume}{6} (\bibinfo{year}{2018}), \bibinfo{pages}{587--604}.
\newblock


\bibitem[Bender et~al\mbox{.}(2021)]%
        {bender2021dangers}
\bibfield{author}{\bibinfo{person}{Emily~M Bender}, \bibinfo{person}{Timnit Gebru}, \bibinfo{person}{Angelina McMillan-Major}, {and} \bibinfo{person}{Shmargaret Shmitchell}.} \bibinfo{year}{2021}\natexlab{}.
\newblock \showarticletitle{On the Dangers of Stochastic Parrots: Can Language Models Be Too Big?}. In \bibinfo{booktitle}{\emph{Proceedings of the 2021 ACM Conference on Fairness, Accountability, and Transparency}}. \bibinfo{pages}{610--623}.
\newblock


\bibitem[Bianchi et~al\mbox{.}(2023)]%
        {bianchi2023easily}
\bibfield{author}{\bibinfo{person}{Federico Bianchi}, \bibinfo{person}{Pratyusha Kalluri}, \bibinfo{person}{Esin Durmus}, \bibinfo{person}{Faisal Ladhak}, \bibinfo{person}{Myra Cheng}, \bibinfo{person}{Debora Nozza}, \bibinfo{person}{Tatsunori Hashimoto}, \bibinfo{person}{Dan Jurafsky}, \bibinfo{person}{James Zou}, {and} \bibinfo{person}{Aylin Caliskan}.} \bibinfo{year}{2023}\natexlab{}.
\newblock \showarticletitle{Easily Accessible Text-to-Image Generation Amplifies Demographic Stereotypes at Large Scale}. In \bibinfo{booktitle}{\emph{Proceedings of the 2023 ACM Conference on Fairness, Accountability, and Transparency}} (Chicago, IL, USA) \emph{(\bibinfo{series}{FAccT '23})}. \bibinfo{publisher}{Association for Computing Machinery}, \bibinfo{address}{New York, NY, USA}, \bibinfo{pages}{1493–1504}.
\newblock
\showISBNx{9798400701924}
\urldef\tempurl%
\url{https://doi.org/10.1145/3593013.3594095}
\showDOI{\tempurl}


\bibitem[Birchall(2021)]%
        {birchall2021radical}
\bibfield{author}{\bibinfo{person}{Clare Birchall}.} \bibinfo{year}{2021}\natexlab{}.
\newblock \bibinfo{booktitle}{\emph{Radical secrecy: The ends of transparency in datafied America}}. Vol.~\bibinfo{volume}{60}.
\newblock \bibinfo{publisher}{U of Minnesota Press}.
\newblock


\bibitem[Bommasani et~al\mbox{.}(2021)]%
        {bommasani2021opportunities}
\bibfield{author}{\bibinfo{person}{Rishi Bommasani}, \bibinfo{person}{Drew~A. Hudson}, \bibinfo{person}{Ehsan Adeli}, \bibinfo{person}{Russ Altman}, \bibinfo{person}{Simran Arora}, \bibinfo{person}{Sydney von Arx}, \bibinfo{person}{Michael~S. Bernstein}, \bibinfo{person}{Jeannette Bohg}, \bibinfo{person}{Antoine Bosselut}, \bibinfo{person}{Emma Brunskill}, \bibinfo{person}{Erik Brynjolfsson}, \bibinfo{person}{Shyamal Buch}, \bibinfo{person}{Dallas Card}, \bibinfo{person}{Rodrigo Castellon}, \bibinfo{person}{Niladri Chatterji}, \bibinfo{person}{Annie Chen}, \bibinfo{person}{Kathleen Creel}, \bibinfo{person}{Jared~Quincy Davis}, \bibinfo{person}{Dorottya Demszky}, \bibinfo{person}{Chris Donahue}, \bibinfo{person}{Moussa Doumbouya}, \bibinfo{person}{Esin Durmus}, \bibinfo{person}{Stefano Ermon}, \bibinfo{person}{John Etchemendy}, \bibinfo{person}{Kawin Ethayarajh}, \bibinfo{person}{Li Fei-Fei}, \bibinfo{person}{Chelsea Finn}, \bibinfo{person}{Trevor Gale}, \bibinfo{person}{Lauren Gillespie},
  \bibinfo{person}{Karan Goel}, \bibinfo{person}{Noah Goodman}, \bibinfo{person}{Shelby Grossman}, \bibinfo{person}{Neel Guha}, \bibinfo{person}{Tatsunori Hashimoto}, \bibinfo{person}{Peter Henderson}, \bibinfo{person}{John Hewitt}, \bibinfo{person}{Daniel~E. Ho}, \bibinfo{person}{Jenny Hong}, \bibinfo{person}{Kyle Hsu}, \bibinfo{person}{Jing Huang}, \bibinfo{person}{Thomas Icard}, \bibinfo{person}{Saahil Jain}, \bibinfo{person}{Dan Jurafsky}, \bibinfo{person}{Pratyusha Kalluri}, \bibinfo{person}{Siddharth Karamcheti}, \bibinfo{person}{Geoff Keeling}, \bibinfo{person}{Fereshte Khani}, \bibinfo{person}{Omar Khattab}, \bibinfo{person}{Pang~Wei Koh}, \bibinfo{person}{Mark Krass}, \bibinfo{person}{Ranjay Krishna}, \bibinfo{person}{Rohith Kuditipudi}, \bibinfo{person}{Ananya Kumar}, \bibinfo{person}{Faisal Ladhak}, \bibinfo{person}{Mina Lee}, \bibinfo{person}{Tony Lee}, \bibinfo{person}{Jure Leskovec}, \bibinfo{person}{Isabelle Levent}, \bibinfo{person}{Xiang~Lisa Li}, \bibinfo{person}{Xuechen Li},
  \bibinfo{person}{Tengyu Ma}, \bibinfo{person}{Ali Malik}, \bibinfo{person}{Christopher~D. Manning}, \bibinfo{person}{Suvir Mirchandani}, \bibinfo{person}{Eric Mitchell}, \bibinfo{person}{Zanele Munyikwa}, \bibinfo{person}{Suraj Nair}, \bibinfo{person}{Avanika Narayan}, \bibinfo{person}{Deepak Narayanan}, \bibinfo{person}{Ben Newman}, \bibinfo{person}{Allen Nie}, \bibinfo{person}{Juan~Carlos Niebles}, \bibinfo{person}{Hamed Nilforoshan}, \bibinfo{person}{Julian Nyarko}, \bibinfo{person}{Giray Ogut}, \bibinfo{person}{Laurel Orr}, \bibinfo{person}{Isabel Papadimitriou}, \bibinfo{person}{Joon~Sung Park}, \bibinfo{person}{Chris Piech}, \bibinfo{person}{Eva Portelance}, \bibinfo{person}{Christopher Potts}, \bibinfo{person}{Aditi Raghunathan}, \bibinfo{person}{Rob Reich}, \bibinfo{person}{Hongyu Ren}, \bibinfo{person}{Frieda Rong}, \bibinfo{person}{Yusuf Roohani}, \bibinfo{person}{Camilo Ruiz}, \bibinfo{person}{Jack Ryan}, \bibinfo{person}{Christopher Ré}, \bibinfo{person}{Dorsa Sadigh}, \bibinfo{person}{Shiori
  Sagawa}, \bibinfo{person}{Keshav Santhanam}, \bibinfo{person}{Andy Shih}, \bibinfo{person}{Krishnan Srinivasan}, \bibinfo{person}{Alex Tamkin}, \bibinfo{person}{Rohan Taori}, \bibinfo{person}{Armin~W. Thomas}, \bibinfo{person}{Florian Tramèr}, \bibinfo{person}{Rose~E. Wang}, \bibinfo{person}{William Wang}, \bibinfo{person}{Bohan Wu}, \bibinfo{person}{Jiajun Wu}, \bibinfo{person}{Yuhuai Wu}, \bibinfo{person}{Sang~Michael Xie}, \bibinfo{person}{Michihiro Yasunaga}, \bibinfo{person}{Jiaxuan You}, \bibinfo{person}{Matei Zaharia}, \bibinfo{person}{Michael Zhang}, \bibinfo{person}{Tianyi Zhang}, \bibinfo{person}{Xikun Zhang}, \bibinfo{person}{Yuhui Zhang}, \bibinfo{person}{Lucia Zheng}, \bibinfo{person}{Kaitlyn Zhou}, {and} \bibinfo{person}{Percy Liang}.} \bibinfo{year}{2021}\natexlab{}.
\newblock \showarticletitle{On the Opportunities and Risks of Foundation Models}.
\newblock \bibinfo{journal}{\emph{arXiv preprint arXiv:2108.07258}} (\bibinfo{year}{2021}).
\newblock


\bibitem[Bommasani et~al\mbox{.}(2023a)]%
        {bommasani2023foundation}
\bibfield{author}{\bibinfo{person}{Rishi Bommasani}, \bibinfo{person}{Kevin Klyman}, \bibinfo{person}{Shayne Longpre}, \bibinfo{person}{Sayash Kapoor}, \bibinfo{person}{Nestor Maslej}, \bibinfo{person}{Betty Xiong}, \bibinfo{person}{Daniel Zhang}, {and} \bibinfo{person}{Percy Liang}.} \bibinfo{year}{2023}\natexlab{a}.
\newblock \bibinfo{title}{The Foundation Model Transparency Index}.
\newblock
\newblock
\showeprint[arxiv]{2310.12941}~[cs.LG]


\bibitem[Bommasani et~al\mbox{.}(2023b)]%
        {bommasani2023ecosystem}
\bibfield{author}{\bibinfo{person}{Rishi Bommasani}, \bibinfo{person}{Dilara Soylu}, \bibinfo{person}{Thomas Liao}, \bibinfo{person}{Kathleen~A. Creel}, {and} \bibinfo{person}{Percy Liang}.} \bibinfo{year}{2023}\natexlab{b}.
\newblock \showarticletitle{Ecosystem Graphs: The Social Footprint of Foundation Models}.
\newblock \bibinfo{journal}{\emph{ArXiv}}  \bibinfo{volume}{abs/2303.15772} (\bibinfo{year}{2023}).
\newblock
\urldef\tempurl%
\url{https://api.semanticscholar.org/CorpusID:257771875}
\showURL{%
\tempurl}


\bibitem[Bommasani et~al\mbox{.}(2023c)]%
        {bommasani2023transparency}
\bibfield{author}{\bibinfo{person}{Rishi Bommasani}, \bibinfo{person}{Daniel Zhang}, \bibinfo{person}{Tony Lee}, {and} \bibinfo{person}{Percy Liang}.} \bibinfo{year}{2023}\natexlab{c}.
\newblock \showarticletitle{Improving Transparency in AI Language Models: A Holistic Evaluation}.
\newblock \bibinfo{journal}{\emph{Foundation Model Issue Brief Series}} (\bibinfo{year}{2023}).
\newblock
\urldef\tempurl%
\url{https://hai.stanford.edu/foundation-model-issue-brief-series}
\showURL{%
\tempurl}


\bibitem[Boyd(2016)]%
        {boyd2016algorithmic}
\bibfield{author}{\bibinfo{person}{Danah Boyd}.} \bibinfo{year}{2016}\natexlab{}.
\newblock \bibinfo{title}{Algorithmic Accountability and Transparency}.
\newblock \bibinfo{howpublished}{Open Transcripts}.
\newblock
\urldef\tempurl%
\url{http://opentranscripts.org/transcript/danah-boyd-algorithmic-accountability-transparency/}
\showURL{%
\tempurl}
\newblock
\shownote{Presented by danah boyd in Algorithmic Accountability and Transparency in the Digital Economy}.


\bibitem[Bubeck et~al\mbox{.}(2023)]%
        {bubeck2023sparks}
\bibfield{author}{\bibinfo{person}{Sébastien Bubeck}, \bibinfo{person}{Varun Chandrasekaran}, \bibinfo{person}{Ronen Eldan}, \bibinfo{person}{Johannes Gehrke}, \bibinfo{person}{Eric Horvitz}, \bibinfo{person}{Ece Kamar}, \bibinfo{person}{Peter Lee}, \bibinfo{person}{Yin~Tat Lee}, \bibinfo{person}{Yuanzhi Li}, \bibinfo{person}{Scott Lundberg}, \bibinfo{person}{Harsha Nori}, \bibinfo{person}{Hamid Palangi}, \bibinfo{person}{Marco~Tulio Ribeiro}, {and} \bibinfo{person}{Yi Zhang}.} \bibinfo{year}{2023}\natexlab{}.
\newblock \bibinfo{title}{Sparks of Artificial General Intelligence: Early experiments with GPT-4}.
\newblock
\newblock
\showeprint[arxiv]{2303.12712}~[cs.CL]


\bibitem[Budish et~al\mbox{.}(2016)]%
        {budish2016transparency}
\bibfield{author}{\bibinfo{person}{Ryan Budish}, \bibinfo{person}{Liz Woolery}, {and} \bibinfo{person}{Kevin Bankston}.} \bibinfo{year}{2016}\natexlab{}.
\newblock \bibinfo{title}{The Transparency Reporting Toolkit: Survey \& Best Practice Memos for Reporting on U.S. Government Requests for User Information}.
\newblock \bibinfo{howpublished}{\url{https://www.newamerica.org/oti/policy-papers/the-transparency-reporting-toolkit/}}.
\newblock


\bibitem[Commission(2022)]%
        {dsa2022}
\bibfield{author}{\bibinfo{person}{European Commission}.} \bibinfo{year}{2022}\natexlab{}.
\newblock \showarticletitle{The Digital Services Act: ensuring a safe and accountable online environment}.
\newblock \bibinfo{journal}{\emph{European Commission}} (\bibinfo{year}{2022}).
\newblock
\urldef\tempurl%
\url{https://commission.europa.eu/strategy-and-policy/priorities-2019-2024/europe-fit-digital-age/digital-services-act-ensuring-safe-and-accountable-online-environment_en}
\showURL{%
\tempurl}


\bibitem[Congress(2023)]%
        {us2023fmta}
\bibfield{author}{\bibinfo{person}{United~States Congress}.} \bibinfo{year}{2023}\natexlab{}.
\newblock \bibinfo{title}{AI Foundation Model Transparency Act}.
\newblock
\newblock
\urldef\tempurl%
\url{https://beyer.house.gov/uploadedfiles/ai_foundation_model_transparency_act_text_118.pdf}
\showURL{%
\tempurl}


\bibitem[Crawford(2021)]%
        {crawford2021atlas}
\bibfield{author}{\bibinfo{person}{Kate Crawford}.} \bibinfo{year}{2021}\natexlab{}.
\newblock \bibinfo{booktitle}{\emph{The atlas of AI: Power, politics, and the planetary costs of artificial intelligence}}.
\newblock \bibinfo{publisher}{Yale University Press}.
\newblock


\bibitem[Crocker et~al\mbox{.}(2019)]%
        {eff2019whyb}
\bibfield{author}{\bibinfo{person}{Andrew Crocker}, \bibinfo{person}{Gennie Gebhart}, \bibinfo{person}{Aaron Mackey}, \bibinfo{person}{Kurt Opsahl}, \bibinfo{person}{Hayley Tsukayama}, \bibinfo{person}{Jamie~Lee Williams}, {and} \bibinfo{person}{Jillian~C. York}.} \bibinfo{year}{2019}\natexlab{}.
\newblock \bibinfo{title}{Who Has Your Back?}
\newblock
\newblock
\urldef\tempurl%
\url{https://www.eff.org/files/2019/06/11/whyb_2019_report.pdf}
\showURL{%
\tempurl}


\bibitem[Edelson et~al\mbox{.}(2021)]%
        {edelson2021universal}
\bibfield{author}{\bibinfo{person}{Laura Edelson}, \bibinfo{person}{Jason Chuang}, \bibinfo{person}{Erika Franklin~Fowler}, \bibinfo{person}{Michael Franz}, {and} \bibinfo{person}{Travis~N. Ridout}.} \bibinfo{year}{2021}\natexlab{}.
\newblock \showarticletitle{Universal Digital Ad Transparency}. In \bibinfo{booktitle}{\emph{TPRC49: The 49th Research Conference on Communication, Information and Internet Policy}}.
\newblock
\newblock
\shownote{Available at SSRN: \url{https://ssrn.com/abstract=3898214} or \url{http://dx.doi.org/10.2139/ssrn.3898214}}.


\bibitem[{European Commission}(2023)]%
        {ec2023consultation}
\bibfield{author}{\bibinfo{person}{{European Commission}}.} \bibinfo{year}{2023}\natexlab{}.
\newblock \bibinfo{title}{Commission launches public consultation on the Implementing Regulation on transparency reporting under the DSA}.
\newblock
\newblock
\urldef\tempurl%
\url{https://digital-strategy.ec.europa.eu/en/news/commission-launches-public-consultation-implementing-regulation-transparency-reporting-under-dsa}
\showURL{%
\tempurl}


\bibitem[{European Council}(2024)]%
        {eu2023aiact}
\bibfield{author}{\bibinfo{person}{{European Council}}.} \bibinfo{year}{2024}\natexlab{}.
\newblock \bibinfo{title}{Proposal for a Regulation of the European Parliament and of the Council laying down harmonised rules on artificial intelligence (Artificial Intelligence Act) and amending certain Union legislative acts}.
\newblock
\newblock
\urldef\tempurl%
\url{https://data.consilium.europa.eu/doc/document/ST-5662-2024-INIT/en/pdf}
\showURL{%
\tempurl}


\bibitem[{Facebook}(2023)]%
        {facebook2023facebook}
\bibfield{author}{\bibinfo{person}{{Facebook}}.} \bibinfo{year}{2023}\natexlab{}.
\newblock \bibinfo{title}{Facebook Transparent Reports}.
\newblock
\newblock
\urldef\tempurl%
\url{https://transparency.fb.com/reports/}
\showURL{%
\tempurl}


\bibitem[Food and Administration(2018)]%
        {faersqa}
\bibfield{author}{\bibinfo{person}{U.S. Food} {and} \bibinfo{person}{Drug Administration}.} \bibinfo{year}{2018}\natexlab{}.
\newblock \bibinfo{title}{Questions and Answers on FDA's Adverse Event Reporting System (FAERS)}.
\newblock \bibinfo{howpublished}{\url{https://www.fda.gov/drugs/surveillance/questions-and-answers-fdas-adverse-event-reporting-system-faers}}.
\newblock


\bibitem[Food and Administration(2021)]%
        {faers}
\bibfield{author}{\bibinfo{person}{U.S. Food} {and} \bibinfo{person}{Drug Administration}.} \bibinfo{year}{2021}\natexlab{}.
\newblock \bibinfo{title}{FDA Adverse Event Reporting System (FAERS): Latest Quartely Data Files}.
\newblock \bibinfo{howpublished}{\url{https://catalog.data.gov/dataset/fda-adverse-event-reporting-system-faers-latest-quartely-data-files}}.
\newblock


\bibitem[Food and Administration(2023)]%
        {dashboard}
\bibfield{author}{\bibinfo{person}{U.S. Food} {and} \bibinfo{person}{Drug Administration}.} \bibinfo{year}{2023}\natexlab{}.
\newblock \bibinfo{title}{FDA Adverse Event Reporting System (FAERS) Public Dashboard}.
\newblock \bibinfo{howpublished}{\url{https://www.fda.gov/drugs/questions-and-answers-fdas-adverse-event-reporting-system-faers/fda-adverse-event-reporting-system-faers-public-dashboard}}.
\newblock


\bibitem[Fourrier et~al\mbox{.}(2023)]%
        {fourrier2023whats}
\bibfield{author}{\bibinfo{person}{Clémentine Fourrier}, \bibinfo{person}{Nathan Habib}, \bibinfo{person}{Julien Launay}, {and} \bibinfo{person}{Thomas Wolf}.} \bibinfo{year}{2023}\natexlab{}.
\newblock \bibinfo{title}{What's going on with the Open LLM Leaderboard?}
\newblock
\newblock
\urldef\tempurl%
\url{https://huggingface.co/blog/evaluating-mmlu-leaderboard}
\showURL{%
\tempurl}


\bibitem[Gebru et~al\mbox{.}(2021)]%
        {gebru2021datasheets}
\bibfield{author}{\bibinfo{person}{Timnit Gebru}, \bibinfo{person}{Jamie Morgenstern}, \bibinfo{person}{Briana Vecchione}, \bibinfo{person}{Jennifer~Wortman Vaughan}, \bibinfo{person}{Hanna Wallach}, \bibinfo{person}{Hal~Daum{\'e} Iii}, {and} \bibinfo{person}{Kate Crawford}.} \bibinfo{year}{2021}\natexlab{}.
\newblock \showarticletitle{Datasheets for datasets}.
\newblock \bibinfo{journal}{\emph{Commun. ACM}} \bibinfo{volume}{64}, \bibinfo{number}{12} (\bibinfo{year}{2021}), \bibinfo{pages}{86--92}.
\newblock


\bibitem[Gebru et~al\mbox{.}(2018)]%
        {gebru2018datasheets}
\bibfield{author}{\bibinfo{person}{Timnit Gebru}, \bibinfo{person}{Jamie Morgenstern}, \bibinfo{person}{Briana Vecchione}, \bibinfo{person}{Jennifer~Wortman Vaughan}, \bibinfo{person}{Hanna Wallach}, \bibinfo{person}{Hal~Daumé Ill}, {and} \bibinfo{person}{Kate Crawford}.} \bibinfo{year}{2018}\natexlab{}.
\newblock \showarticletitle{Datasheets for Datasets}.
\newblock \bibinfo{journal}{\emph{arXiv preprint arXiv:1803.09010}} (\bibinfo{year}{2018}).
\newblock


\bibitem[Ghosh and Faxon(2023)]%
        {doi:10.1177/20539517231164119}
\bibfield{author}{\bibinfo{person}{Ritwick Ghosh} {and} \bibinfo{person}{Hilary~Oliva Faxon}.} \bibinfo{year}{2023}\natexlab{}.
\newblock \showarticletitle{Smart corruption: Satirical strategies for gaming accountability}.
\newblock \bibinfo{journal}{\emph{Big Data \& Society}} \bibinfo{volume}{10}, \bibinfo{number}{1} (\bibinfo{year}{2023}), \bibinfo{pages}{20539517231164119}.
\newblock
\urldef\tempurl%
\url{https://doi.org/10.1177/20539517231164119}
\showDOI{\tempurl}
\showeprint{https://doi.org/10.1177/20539517231164119}


\bibitem[Gorwa and Ash(2020)]%
        {gorwa_ash_2020}
\bibfield{author}{\bibinfo{person}{Robert Gorwa} {and} \bibinfo{person}{Timothy~Garton Ash}.} \bibinfo{year}{2020}\natexlab{}.
\newblock \bibinfo{booktitle}{\emph{Democratic Transparency in the Platform Society}}.
\newblock \bibinfo{publisher}{Cambridge University Press}, \bibinfo{pages}{286–312}.
\newblock


\bibitem[Gray and Suri(2019)]%
        {gray2019ghost}
\bibfield{author}{\bibinfo{person}{Mary~L Gray} {and} \bibinfo{person}{Siddharth Suri}.} \bibinfo{year}{2019}\natexlab{}.
\newblock \bibinfo{booktitle}{\emph{Ghost work: How to stop Silicon Valley from building a new global underclass}}.
\newblock \bibinfo{publisher}{Eamon Dolan Books}.
\newblock


\bibitem[Greenwald(2013)]%
        {greenwald2013nsa}
\bibfield{author}{\bibinfo{person}{Glenn Greenwald}.} \bibinfo{year}{2013}\natexlab{}.
\newblock \bibinfo{title}{NSA collecting phone records of millions of Verizon customers daily}.
\newblock \bibinfo{howpublished}{\url{https://www.theguardian.com/world/2013/jun/06/nsa-phone-records-verizon-court-order}}.
\newblock


\bibitem[{Group of Seven}(2023)]%
        {g72023vc}
\bibfield{author}{\bibinfo{person}{{Group of Seven}}.} \bibinfo{year}{2023}\natexlab{}.
\newblock \bibinfo{title}{Hiroshima Process International Code of Conduct for Organizations Developing Advanced AI Syste}.
\newblock
\newblock
\urldef\tempurl%
\url{https://www.mofa.go.jp/files/100573473.pdf}
\showURL{%
\tempurl}


\bibitem[Guha et~al\mbox{.}(2023)]%
        {guha2023ai}
\bibfield{author}{\bibinfo{person}{Neel Guha}, \bibinfo{person}{Christie~M. Lawrence}, \bibinfo{person}{Lindsey~A. Gailmard}, \bibinfo{person}{Kit~T. Rodolfa}, \bibinfo{person}{Faiz Surani}, \bibinfo{person}{Rishi Bommasani}, \bibinfo{person}{Inioluwa~Deborah Raji}, \bibinfo{person}{Mariano-Florentino Cu{\'e}llar}, \bibinfo{person}{Colleen Honigsberg}, \bibinfo{person}{Percy Liang}, {and} \bibinfo{person}{Daniel~E. Ho}.} \bibinfo{year}{2023}\natexlab{}.
\newblock \showarticletitle{AI Regulation Has Its Own Alignment Problem: The Technical and Institutional Feasibility of Disclosure, Registration, Licensing, and Auditing}.
\newblock \bibinfo{journal}{\emph{George Washington Law Review, Symposium on Legally Disruptive Emerging Technologies}} (\bibinfo{year}{2023}).
\newblock


\bibitem[Han(2015)]%
        {han2015transparency}
\bibfield{author}{\bibinfo{person}{Byung-Chul Han}.} \bibinfo{year}{2015}\natexlab{}.
\newblock \bibinfo{booktitle}{\emph{The transparency society}}.
\newblock \bibinfo{publisher}{Stanford University Press}.
\newblock


\bibitem[Hao and Seetharaman(2023)]%
        {hao2023cleaning}
\bibfield{author}{\bibinfo{person}{Karen Hao} {and} \bibinfo{person}{Deepa Seetharaman}.} \bibinfo{year}{2023}\natexlab{}.
\newblock \showarticletitle{Cleaning Up ChatGPT Takes Heavy Toll on Human Workers}.
\newblock \bibinfo{journal}{\emph{The Wall Street Journal}} (\bibinfo{date}{24 July} \bibinfo{year}{2023}).
\newblock
\urldef\tempurl%
\url{https://www.wsj.com/articles/chatgpt-openai-content-abusive-sexually-explicit-harassment-kenya-workers-on-human-workers-cf191483}
\showURL{%
\tempurl}
\newblock
\shownote{Photographs by Natalia Jidovanu}.


\bibitem[Hartzog(2023)]%
        {hartzog2023oversight}
\bibfield{author}{\bibinfo{person}{Woodrow Hartzog}.} \bibinfo{year}{2023}\natexlab{}.
\newblock \bibinfo{title}{Oversight of A.I.: Legislating on Artificial Intelligence}.
\newblock \bibinfo{howpublished}{Prepared Testimony and Statement for the Record before the U.S. Senate Committee on the Judiciary, Subcommittee on Privacy, Technology, and the Law}.
\newblock
\urldef\tempurl%
\url{https://www.judiciary.senate.gov/imo/media/doc/2023-09-12_pm_-_testimony_-_hartzog.pdf}
\showURL{%
\tempurl}


\bibitem[Hendrycks et~al\mbox{.}(2021)]%
        {hendrycks2021measuring}
\bibfield{author}{\bibinfo{person}{Dan Hendrycks}, \bibinfo{person}{Collin Burns}, \bibinfo{person}{Steven Basart}, \bibinfo{person}{Andy Zou}, \bibinfo{person}{Mantas Mazeika}, \bibinfo{person}{Dawn Song}, {and} \bibinfo{person}{Jacob Steinhardt}.} \bibinfo{year}{2021}\natexlab{}.
\newblock \showarticletitle{Measuring massive multitask language understanding}. In \bibinfo{booktitle}{\emph{International Conference on Learning Representations (ICLR)}}.
\newblock


\bibitem[House(2023)]%
        {us2023vc}
\bibfield{author}{\bibinfo{person}{The~White House}.} \bibinfo{year}{2023}\natexlab{}.
\newblock \bibinfo{title}{Ensuring Safe, Secure, and Trustworthy AI}.
\newblock
\newblock
\urldef\tempurl%
\url{https://www.whitehouse.gov/wp-content/uploads/2023/07/Ensuring-Safe-Secure-and-Trustworthy-AI.pdf}
\showURL{%
\tempurl}


\bibitem[{Innovation, Science and Economic Development Canada}(2023)]%
        {canada2023vc}
\bibfield{author}{\bibinfo{person}{{Innovation, Science and Economic Development Canada}}.} \bibinfo{year}{2023}\natexlab{}.
\newblock \bibinfo{title}{Voluntary Code of Conduct on the Responsible Development and Management of Advanced Generative AI Systems}.
\newblock
\newblock
\urldef\tempurl%
\url{https://ised-isde.canada.ca/site/ised/en/voluntary-code-conduct-responsible-development-and-management-advanced-generative-ai-systems}
\showURL{%
\tempurl}


\bibitem[Kapoor and Narayanan(2023)]%
        {kapoor2023licensing}
\bibfield{author}{\bibinfo{person}{Sayash Kapoor} {and} \bibinfo{person}{Arvind Narayanan}.} \bibinfo{year}{2023}\natexlab{}.
\newblock \bibinfo{booktitle}{\emph{Licensing is neither feasible nor effective for addressing AI risks}}.
\newblock
\urldef\tempurl%
\url{https://www.aisnakeoil.com/p/licensing-is-neither-feasible-nor}
\showURL{%
\tempurl}


\bibitem[Keller(2021)]%
        {keller2021humility}
\bibfield{author}{\bibinfo{person}{Daphne Keller}.} \bibinfo{year}{2021}\natexlab{}.
\newblock \bibinfo{title}{Some Humility About Transparency}.
\newblock \bibinfo{howpublished}{\url{https://cyberlaw.stanford.edu/blog/2021/03/some-humility-about-transparency}}.
\newblock


\bibitem[Keller(2022)]%
        {keller2022platform}
\bibfield{author}{\bibinfo{person}{Daphne Keller}.} \bibinfo{year}{2022}\natexlab{}.
\newblock \bibinfo{booktitle}{\emph{Hearing on Platform Transparency: Understanding the Impact of Social Media}}.
\newblock \bibinfo{type}{{T}echnical {R}eport}. \bibinfo{institution}{United States Senate Committee on the Judiciary, Subcommittee on Privacy, Technology and the Law}.
\newblock
\urldef\tempurl%
\url{https://www.judiciary.senate.gov/imo/media/doc/Keller%20Testimony1.pdf}
\showURL{%
\tempurl}
\newblock
\shownote{Statement of Daphne Keller, Stanford University Cyber Policy Center}.


\bibitem[Kessel(2016)]%
        {kessel2016advancing}
\bibfield{author}{\bibinfo{person}{Jeremy Kessel}.} \bibinfo{year}{2016}\natexlab{}.
\newblock \bibinfo{title}{Advancing \#transparency with more insightful data}.
\newblock \bibinfo{howpublished}{\url{https://blog.twitter.com/official/en_us/a/2016/advancing-transparency-with-more-insightful-data.html}}.
\newblock


\bibitem[Kumar(2018)]%
        {faersimplication}
\bibfield{author}{\bibinfo{person}{Atul Kumar}.} \bibinfo{year}{2018}\natexlab{}.
\newblock \bibinfo{title}{The Newly Available FAERS Public Dashboard: Implications for Health Care Professionals}.
\newblock
\newblock
Issue 2.


\bibitem[Lazar(2023)]%
        {lazar2023algorithmiccity}
\bibfield{author}{\bibinfo{person}{Seth Lazar}.} \bibinfo{year}{2023}\natexlab{}.
\newblock \showarticletitle{Governing the Algorithmic City}.
\newblock \bibinfo{journal}{\emph{Tanner Lectures}} (\bibinfo{year}{2023}).
\newblock
\urldef\tempurl%
\url{https://write.as/sethlazar/}
\showURL{%
\tempurl}


\bibitem[Le~Scao et~al\mbox{.}(2022)]%
        {scao2022bloom}
\bibfield{author}{\bibinfo{person}{Teven Le~Scao}, \bibinfo{person}{Angela Fan}, \bibinfo{person}{Christopher Akiki}, \bibinfo{person}{Ellie Pavlick}, \bibinfo{person}{Suzana Ilić}, \bibinfo{person}{Daniel Hesslow}, \bibinfo{person}{Roman Castagné}, \bibinfo{person}{Alexandra~Sasha Luccioni}, \bibinfo{person}{François Yvon}, \bibinfo{person}{Matthias Gallé}, \bibinfo{person}{Jonathan Tow}, \bibinfo{person}{Alexander~M. Rush}, \bibinfo{person}{Stella Biderman}, \bibinfo{person}{Albert Webson}, \bibinfo{person}{Pawan~Sasanka Ammanamanchi}, \bibinfo{person}{Thomas Wang}, \bibinfo{person}{Benoît Sagot}, \bibinfo{person}{Niklas Muennighoff}, \bibinfo{person}{Albert~Villanova del Moral}, \bibinfo{person}{Olatunji Ruwase}, \bibinfo{person}{Rachel Bawden}, \bibinfo{person}{Stas Bekman}, \bibinfo{person}{Angelina McMillan-Major}, \bibinfo{person}{Iz Beltagy}, \bibinfo{person}{Huu Nguyen}, \bibinfo{person}{Lucile Saulnier}, \bibinfo{person}{Samson Tan}, \bibinfo{person}{Pedro~Ortiz Suarez}, \bibinfo{person}{Victor
  Sanh}, \bibinfo{person}{Hugo Laurençon}, \bibinfo{person}{Yacine Jernite}, \bibinfo{person}{Julien Launay}, \bibinfo{person}{Margaret Mitchell}, \bibinfo{person}{Colin Raffel}, \bibinfo{person}{Aaron Gokaslan}, \bibinfo{person}{Adi Simhi}, \bibinfo{person}{Aitor Soroa}, \bibinfo{person}{Alham~Fikri Aji}, \bibinfo{person}{Amit Alfassy}, \bibinfo{person}{Anna Rogers}, \bibinfo{person}{Ariel~Kreisberg Nitzav}, \bibinfo{person}{Canwen Xu}, \bibinfo{person}{Chenghao Mou}, \bibinfo{person}{Chris Emezue}, \bibinfo{person}{Christopher Klamm}, \bibinfo{person}{Colin Leong}, \bibinfo{person}{Daniel van Strien}, \bibinfo{person}{David~Ifeoluwa Adelani}, \bibinfo{person}{Dragomir Radev}, \bibinfo{person}{Eduardo~González Ponferrada}, \bibinfo{person}{Efrat Levkovizh}, \bibinfo{person}{Ethan Kim}, \bibinfo{person}{Eyal~Bar Natan}, \bibinfo{person}{Francesco De~Toni}, \bibinfo{person}{Gérard Dupont}, \bibinfo{person}{Germán Kruszewski}, \bibinfo{person}{Giada Pistilli}, \bibinfo{person}{Hady Elsahar},
  \bibinfo{person}{Hamza Benyamina}, \bibinfo{person}{Hieu Tran}, \bibinfo{person}{Ian Yu}, \bibinfo{person}{Idris Abdulmumin}, \bibinfo{person}{Isaac Johnson}, \bibinfo{person}{Itziar Gonzalez-Dios}, \bibinfo{person}{Javier de~la Rosa}, \bibinfo{person}{Jenny Chim}, \bibinfo{person}{Jesse Dodge}, \bibinfo{person}{Jian Zhu}, \bibinfo{person}{Jonathan Chang}, \bibinfo{person}{Jörg Frohberg}, \bibinfo{person}{Joseph Tobing}, \bibinfo{person}{Joydeep Bhattacharjee}, \bibinfo{person}{Khalid Almubarak}, \bibinfo{person}{Kimbo Chen}, \bibinfo{person}{Kyle Lo}, \bibinfo{person}{Leandro Von~Werra}, \bibinfo{person}{Leon Weber}, \bibinfo{person}{Long Phan}, \bibinfo{person}{Loubna~Ben allal}, \bibinfo{person}{Ludovic Tanguy}, \bibinfo{person}{Manan Dey}, \bibinfo{person}{Manuel~Romero Muñoz}, \bibinfo{person}{Maraim Masoud}, \bibinfo{person}{María Grandury}, \bibinfo{person}{Mario Šaško}, \bibinfo{person}{Max Huang}, \bibinfo{person}{Maximin Coavoux}, \bibinfo{person}{Mayank Singh}, \bibinfo{person}{Mike
  Tian-Jian Jiang}, \bibinfo{person}{Minh~Chien Vu}, \bibinfo{person}{Mohammad~A. Jauhar}, \bibinfo{person}{Mustafa Ghaleb}, \bibinfo{person}{Nishant Subramani}, \bibinfo{person}{Nora Kassner}, \bibinfo{person}{Nurulaqilla Khamis}, \bibinfo{person}{Olivier Nguyen}, \bibinfo{person}{Omar Espejel}, \bibinfo{person}{Ona de Gibert}, \bibinfo{person}{Paulo Villegas}, \bibinfo{person}{Peter Henderson}, \bibinfo{person}{Pierre Colombo}, \bibinfo{person}{Priscilla Amuok}, \bibinfo{person}{Quentin Lhoest}, \bibinfo{person}{Rheza Harliman}, \bibinfo{person}{Rishi Bommasani}, \bibinfo{person}{Roberto~Luis López}, \bibinfo{person}{Rui Ribeiro}, \bibinfo{person}{Salomey Osei}, \bibinfo{person}{Sampo Pyysalo}, \bibinfo{person}{Sebastian Nagel}, \bibinfo{person}{Shamik Bose}, \bibinfo{person}{Shamsuddeen~Hassan Muhammad}, \bibinfo{person}{Shanya Sharma}, \bibinfo{person}{Shayne Longpre}, \bibinfo{person}{Somaieh Nikpoor}, \bibinfo{person}{Stanislav Silberberg}, \bibinfo{person}{Suhas Pai}, \bibinfo{person}{Sydney Zink},
  \bibinfo{person}{Tiago~Timponi Torrent}, \bibinfo{person}{Timo Schick}, \bibinfo{person}{Tristan Thrush}, \bibinfo{person}{Valentin Danchev}, \bibinfo{person}{Vassilina Nikoulina}, \bibinfo{person}{Veronika Laippala}, \bibinfo{person}{Violette Lepercq}, \bibinfo{person}{Vrinda Prabhu}, \bibinfo{person}{Zaid Alyafeai}, \bibinfo{person}{Zeerak Talat}, \bibinfo{person}{Arun Raja}, \bibinfo{person}{Benjamin Heinzerling}, \bibinfo{person}{Chenglei Si}, \bibinfo{person}{Elizabeth Salesky}, \bibinfo{person}{Sabrina~J. Mielke}, \bibinfo{person}{Wilson~Y. Lee}, \bibinfo{person}{Abheesht Sharma}, \bibinfo{person}{Andrea Santilli}, \bibinfo{person}{Antoine Chaffin}, \bibinfo{person}{Arnaud Stiegler}, \bibinfo{person}{Debajyoti Datta}, \bibinfo{person}{Eliza Szczechla}, \bibinfo{person}{Gunjan Chhablani}, \bibinfo{person}{Han Wang}, \bibinfo{person}{Harshit Pandey}, \bibinfo{person}{Hendrik Strobelt}, \bibinfo{person}{Jason~Alan Fries}, \bibinfo{person}{Jos Rozen}, \bibinfo{person}{Leo Gao}, \bibinfo{person}{Lintang
  Sutawika}, \bibinfo{person}{M~Saiful Bari}, \bibinfo{person}{Maged~S. Al-shaibani}, \bibinfo{person}{Matteo Manica}, \bibinfo{person}{Nihal Nayak}, \bibinfo{person}{Ryan Teehan}, \bibinfo{person}{Samuel Albanie}, \bibinfo{person}{Sheng Shen}, \bibinfo{person}{Srulik Ben-David}, \bibinfo{person}{Stephen~H. Bach}, \bibinfo{person}{Taewoon Kim}, \bibinfo{person}{Tali Bers}, \bibinfo{person}{Thibault Fevry}, \bibinfo{person}{Trishala Neeraj}, \bibinfo{person}{Urmish Thakker}, \bibinfo{person}{Vikas Raunak}, \bibinfo{person}{Xiangru Tang}, \bibinfo{person}{Zheng-Xin Yong}, \bibinfo{person}{Zhiqing Sun}, \bibinfo{person}{Shaked Brody}, \bibinfo{person}{Yallow Uri}, \bibinfo{person}{Hadar Tojarieh}, \bibinfo{person}{Adam Roberts}, \bibinfo{person}{Hyung~Won Chung}, \bibinfo{person}{Jaesung Tae}, \bibinfo{person}{Jason Phang}, \bibinfo{person}{Ofir Press}, \bibinfo{person}{Conglong Li}, \bibinfo{person}{Deepak Narayanan}, \bibinfo{person}{Hatim Bourfoune}, \bibinfo{person}{Jared Casper}, \bibinfo{person}{Jeff
  Rasley}, \bibinfo{person}{Max Ryabinin}, \bibinfo{person}{Mayank Mishra}, \bibinfo{person}{Minjia Zhang}, \bibinfo{person}{Mohammad Shoeybi}, \bibinfo{person}{Myriam Peyrounette}, \bibinfo{person}{Nicolas Patry}, \bibinfo{person}{Nouamane Tazi}, \bibinfo{person}{Omar Sanseviero}, \bibinfo{person}{Patrick von Platen}, \bibinfo{person}{Pierre Cornette}, \bibinfo{person}{Pierre~François Lavallée}, \bibinfo{person}{Rémi Lacroix}, \bibinfo{person}{Samyam Rajbhandari}, \bibinfo{person}{Sanchit Gandhi}, \bibinfo{person}{Shaden Smith}, \bibinfo{person}{Stéphane Requena}, \bibinfo{person}{Suraj Patil}, \bibinfo{person}{Tim Dettmers}, \bibinfo{person}{Ahmed Baruwa}, \bibinfo{person}{Amanpreet Singh}, \bibinfo{person}{Anastasia Cheveleva}, \bibinfo{person}{Anne-Laure Ligozat}, \bibinfo{person}{Arjun Subramonian}, \bibinfo{person}{Aurélie Névéol}, \bibinfo{person}{Charles Lovering}, \bibinfo{person}{Dan Garrette}, \bibinfo{person}{Deepak Tunuguntla}, \bibinfo{person}{Ehud Reiter}, \bibinfo{person}{Ekaterina
  Taktasheva}, \bibinfo{person}{Ekaterina Voloshina}, \bibinfo{person}{Eli Bogdanov}, \bibinfo{person}{Genta~Indra Winata}, \bibinfo{person}{Hailey Schoelkopf}, \bibinfo{person}{Jan-Christoph Kalo}, \bibinfo{person}{Jekaterina Novikova}, \bibinfo{person}{Jessica~Zosa Forde}, \bibinfo{person}{Jordan Clive}, \bibinfo{person}{Jungo Kasai}, \bibinfo{person}{Ken Kawamura}, \bibinfo{person}{Liam Hazan}, \bibinfo{person}{Marine Carpuat}, \bibinfo{person}{Miruna Clinciu}, \bibinfo{person}{Najoung Kim}, \bibinfo{person}{Newton Cheng}, \bibinfo{person}{Oleg Serikov}, \bibinfo{person}{Omer Antverg}, \bibinfo{person}{Oskar van~der Wal}, \bibinfo{person}{Rui Zhang}, \bibinfo{person}{Ruochen Zhang}, \bibinfo{person}{Sebastian Gehrmann}, \bibinfo{person}{Shani Pais}, \bibinfo{person}{Tatiana Shavrina}, \bibinfo{person}{Thomas Scialom}, \bibinfo{person}{Tian Yun}, \bibinfo{person}{Tomasz Limisiewicz}, \bibinfo{person}{Verena Rieser}, \bibinfo{person}{Vitaly Protasov}, \bibinfo{person}{Vladislav Mikhailov},
  \bibinfo{person}{Yada Pruksachatkun}, \bibinfo{person}{Yonatan Belinkov}, \bibinfo{person}{Zachary Bamberger}, \bibinfo{person}{Zdeněk Kasner}, \bibinfo{person}{Alice Rueda}, \bibinfo{person}{Amanda Pestana}, \bibinfo{person}{Amir Feizpour}, \bibinfo{person}{Ammar Khan}, \bibinfo{person}{Amy Faranak}, \bibinfo{person}{Ana Santos}, \bibinfo{person}{Anthony Hevia}, \bibinfo{person}{Antigona Unldreaj}, \bibinfo{person}{Arash Aghagol}, \bibinfo{person}{Arezoo Abdollahi}, \bibinfo{person}{Aycha Tammour}, \bibinfo{person}{Azadeh HajiHosseini}, \bibinfo{person}{Bahareh Behroozi}, \bibinfo{person}{Benjamin Ajibade}, \bibinfo{person}{Bharat Saxena}, \bibinfo{person}{Carlos~Muñoz Ferrandis}, \bibinfo{person}{Danish Contractor}, \bibinfo{person}{David Lansky}, \bibinfo{person}{Davis David}, \bibinfo{person}{Douwe Kiela}, \bibinfo{person}{Duong~A. Nguyen}, \bibinfo{person}{Edward Tan}, \bibinfo{person}{Emi Baylor}, \bibinfo{person}{Ezinwanne Ozoani}, \bibinfo{person}{Fatima Mirza}, \bibinfo{person}{Frankline
  Ononiwu}, \bibinfo{person}{Habib Rezanejad}, \bibinfo{person}{Hessie Jones}, \bibinfo{person}{Indrani Bhattacharya}, \bibinfo{person}{Irene Solaiman}, \bibinfo{person}{Irina Sedenko}, \bibinfo{person}{Isar Nejadgholi}, \bibinfo{person}{Jesse Passmore}, \bibinfo{person}{Josh Seltzer}, \bibinfo{person}{Julio~Bonis Sanz}, \bibinfo{person}{Karen Fort}, \bibinfo{person}{Livia Dutra}, \bibinfo{person}{Mairon Samagaio}, \bibinfo{person}{Maraim Elbadri}, \bibinfo{person}{Margot Mieskes}, \bibinfo{person}{Marissa Gerchick}, \bibinfo{person}{Martha Akinlolu}, \bibinfo{person}{Michael McKenna}, \bibinfo{person}{Mike Qiu}, \bibinfo{person}{Muhammed Ghauri}, \bibinfo{person}{Mykola Burynok}, \bibinfo{person}{Nafis Abrar}, \bibinfo{person}{Nazneen Rajani}, \bibinfo{person}{Nour Elkott}, \bibinfo{person}{Nour Fahmy}, \bibinfo{person}{Olanrewaju Samuel}, \bibinfo{person}{Ran An}, \bibinfo{person}{Rasmus Kromann}, \bibinfo{person}{Ryan Hao}, \bibinfo{person}{Samira Alizadeh}, \bibinfo{person}{Sarmad Shubber},
  \bibinfo{person}{Silas Wang}, \bibinfo{person}{Sourav Roy}, \bibinfo{person}{Sylvain Viguier}, \bibinfo{person}{Thanh Le}, \bibinfo{person}{Tobi Oyebade}, \bibinfo{person}{Trieu Le}, \bibinfo{person}{Yoyo Yang}, \bibinfo{person}{Zach Nguyen}, \bibinfo{person}{Abhinav~Ramesh Kashyap}, \bibinfo{person}{Alfredo Palasciano}, \bibinfo{person}{Alison Callahan}, \bibinfo{person}{Anima Shukla}, \bibinfo{person}{Antonio Miranda-Escalada}, \bibinfo{person}{Ayush Singh}, \bibinfo{person}{Benjamin Beilharz}, \bibinfo{person}{Bo Wang}, \bibinfo{person}{Caio Brito}, \bibinfo{person}{Chenxi Zhou}, \bibinfo{person}{Chirag Jain}, \bibinfo{person}{Chuxin Xu}, \bibinfo{person}{Clémentine Fourrier}, \bibinfo{person}{Daniel~León Periñán}, \bibinfo{person}{Daniel Molano}, \bibinfo{person}{Dian Yu}, \bibinfo{person}{Enrique Manjavacas}, \bibinfo{person}{Fabio Barth}, \bibinfo{person}{Florian Fuhrimann}, \bibinfo{person}{Gabriel Altay}, \bibinfo{person}{Giyaseddin Bayrak}, \bibinfo{person}{Gully Burns},
  \bibinfo{person}{Helena~U. Vrabec}, \bibinfo{person}{Imane Bello}, \bibinfo{person}{Ishani Dash}, \bibinfo{person}{Jihyun Kang}, \bibinfo{person}{John Giorgi}, \bibinfo{person}{Jonas Golde}, \bibinfo{person}{Jose~David Posada}, \bibinfo{person}{Karthik~Rangasai Sivaraman}, \bibinfo{person}{Lokesh Bulchandani}, \bibinfo{person}{Lu Liu}, \bibinfo{person}{Luisa Shinzato}, \bibinfo{person}{Madeleine~Hahn de Bykhovetz}, \bibinfo{person}{Maiko Takeuchi}, \bibinfo{person}{Marc Pàmies}, \bibinfo{person}{Maria~A Castillo}, \bibinfo{person}{Marianna Nezhurina}, \bibinfo{person}{Mario Sänger}, \bibinfo{person}{Matthias Samwald}, \bibinfo{person}{Michael Cullan}, \bibinfo{person}{Michael Weinberg}, \bibinfo{person}{Michiel De~Wolf}, \bibinfo{person}{Mina Mihaljcic}, \bibinfo{person}{Minna Liu}, \bibinfo{person}{Moritz Freidank}, \bibinfo{person}{Myungsun Kang}, \bibinfo{person}{Natasha Seelam}, \bibinfo{person}{Nathan Dahlberg}, \bibinfo{person}{Nicholas~Michio Broad}, \bibinfo{person}{Nikolaus Muellner},
  \bibinfo{person}{Pascale Fung}, \bibinfo{person}{Patrick Haller}, \bibinfo{person}{Ramya Chandrasekhar}, \bibinfo{person}{Renata Eisenberg}, \bibinfo{person}{Robert Martin}, \bibinfo{person}{Rodrigo Canalli}, \bibinfo{person}{Rosaline Su}, \bibinfo{person}{Ruisi Su}, \bibinfo{person}{Samuel Cahyawijaya}, \bibinfo{person}{Samuele Garda}, \bibinfo{person}{Shlok~S Deshmukh}, \bibinfo{person}{Shubhanshu Mishra}, \bibinfo{person}{Sid Kiblawi}, \bibinfo{person}{Simon Ott}, \bibinfo{person}{Sinee Sang-aroonsiri}, \bibinfo{person}{Srishti Kumar}, \bibinfo{person}{Stefan Schweter}, \bibinfo{person}{Sushil Bharati}, \bibinfo{person}{Tanmay Laud}, \bibinfo{person}{Théo Gigant}, \bibinfo{person}{Tomoya Kainuma}, \bibinfo{person}{Wojciech Kusa}, \bibinfo{person}{Yanis Labrak}, \bibinfo{person}{Yash~Shailesh Bajaj}, \bibinfo{person}{Yash Venkatraman}, \bibinfo{person}{Yifan Xu}, \bibinfo{person}{Yingxin Xu}, \bibinfo{person}{Yu Xu}, \bibinfo{person}{Zhe Tan}, \bibinfo{person}{Zhongli Xie}, \bibinfo{person}{Zifan Ye},
  \bibinfo{person}{Mathilde Bras}, \bibinfo{person}{Younes Belkada}, {and} \bibinfo{person}{Thomas Wolf}.} \bibinfo{year}{2022}\natexlab{}.
\newblock \showarticletitle{BLOOM: A 176B-Parameter Open-Access Multilingual Language Model}.
\newblock  (\bibinfo{year}{2022}).
\newblock
\urldef\tempurl%
\url{https://doi.org/10.48550/ARXIV.2211.05100}
\showDOI{\tempurl}


\bibitem[Llans{\'o} and Vogus(2021)]%
        {llanso2021transparency}
\bibfield{author}{\bibinfo{person}{Emma Llans{\'o}} {and} \bibinfo{person}{Caitlin Vogus}.} \bibinfo{year}{2021}\natexlab{}.
\newblock \bibinfo{title}{Transparency Reports}.
\newblock \bibinfo{howpublished}{\url{https://cdt.org/wp-content/uploads/2022/01/2021-12-20-FX-Transparency-Framework-brief-Transparency-Reports-final.pdf}}.
\newblock


\bibitem[Luccioni et~al\mbox{.}(2023)]%
        {luccioni2023stable}
\bibfield{author}{\bibinfo{person}{Alexandra~Sasha Luccioni}, \bibinfo{person}{Christopher Akiki}, \bibinfo{person}{Margaret Mitchell}, {and} \bibinfo{person}{Yacine Jernite}.} \bibinfo{year}{2023}\natexlab{}.
\newblock \bibinfo{title}{Stable Bias: Analyzing Societal Representations in Diffusion Models}.
\newblock
\newblock
\showeprint[arxiv]{2303.11408}~[cs.CY]


\bibitem[Luccioni et~al\mbox{.}(2022)]%
        {luccioni2022estimating}
\bibfield{author}{\bibinfo{person}{Alexandra~Sasha Luccioni}, \bibinfo{person}{Sylvain Viguier}, {and} \bibinfo{person}{Anne-Laure Ligozat}.} \bibinfo{year}{2022}\natexlab{}.
\newblock \showarticletitle{Estimating the Carbon Footprint of BLOOM, a 176B Parameter Language Model}.
\newblock \bibinfo{journal}{\emph{ArXiv}}  \bibinfo{volume}{abs/2211.02001} (\bibinfo{year}{2022}).
\newblock
\urldef\tempurl%
\url{https://api.semanticscholar.org/CorpusID:253265387}
\showURL{%
\tempurl}


\bibitem[MacKinnon et~al\mbox{.}(2019)]%
        {rdr2019corpacc}
\bibfield{author}{\bibinfo{person}{Rebecca MacKinnon}, \bibinfo{person}{Amy Brouillette}, \bibinfo{person}{Lisa Gutermuth}, \bibinfo{person}{Laura Reed}, \bibinfo{person}{Nathalie Maréchal}, \bibinfo{person}{Veszna Wessenauer}, \bibinfo{person}{Afef Abrougui}, \bibinfo{person}{Sam Cabral}, \bibinfo{person}{Ilja Sperling}, \bibinfo{person}{Zak Rogoff}, {and} \bibinfo{person}{Eeva Moore}.} \bibinfo{year}{2019}\natexlab{}.
\newblock \bibinfo{title}{2019 RDR Corporate Accountability Index}.
\newblock
\newblock
\urldef\tempurl%
\url{https://rankingdigitalrights.org/index2019/assets/static/download/RDRindex2019report.pdf}
\showURL{%
\tempurl}


\bibitem[Meinhardt et~al\mbox{.}(2023)]%
        {meinhardt2023tracking}
\bibfield{author}{\bibinfo{person}{Caroline Meinhardt}, \bibinfo{person}{Christie~M. Lawrence}, \bibinfo{person}{Lindsey~A. Gailmard}, \bibinfo{person}{Daniel Zhang}, \bibinfo{person}{Rishi Bommasani}, \bibinfo{person}{Rohini Kosoglu}, \bibinfo{person}{Peter Henderson}, \bibinfo{person}{Russell Wald}, {and} \bibinfo{person}{Daniel~E. Ho}.} \bibinfo{year}{2023}\natexlab{}.
\newblock \bibinfo{title}{By the Numbers: Tracking The AI Executive Order}.
\newblock
\newblock
\urldef\tempurl%
\url{https://hai.stanford.edu/news/numbers-tracking-ai-executive-order}
\showURL{%
\tempurl}


\bibitem[Miller(2023)]%
        {miller2023tracking}
\bibfield{author}{\bibinfo{person}{Gabby Miller}.} \bibinfo{year}{2023}\natexlab{}.
\newblock \bibinfo{title}{Tracking the First Digital Services Act Transparency Reports}.
\newblock \bibinfo{howpublished}{\url{https://www.techpolicy.press/tracking-the-first-digital-services-act-transparency-reports/}}.
\newblock


\bibitem[Mitchell et~al\mbox{.}(2018)]%
        {mitchell2018modelcards}
\bibfield{author}{\bibinfo{person}{Margaret Mitchell}, \bibinfo{person}{Simone Wu}, \bibinfo{person}{Andrew Zaldivar}, \bibinfo{person}{Parker Barnes}, \bibinfo{person}{Lucy Vasserman}, \bibinfo{person}{Ben Hutchinson}, \bibinfo{person}{Elena Spitzer}, \bibinfo{person}{Inioluwa~Deborah Raji}, {and} \bibinfo{person}{Timnit Gebru}.} \bibinfo{year}{2018}\natexlab{}.
\newblock \showarticletitle{Model Cards for Model Reporting}.
\newblock \bibinfo{journal}{\emph{Proceedings of the Conference on Fairness, Accountability, and Transparency}} (\bibinfo{year}{2018}).
\newblock


\bibitem[Mittelstadt(2019)]%
        {Mittelstadt2019}
\bibfield{author}{\bibinfo{person}{Brent Mittelstadt}.} \bibinfo{year}{2019}\natexlab{}.
\newblock \showarticletitle{Principles alone cannot guarantee ethical AI}.
\newblock \bibinfo{journal}{\emph{Nature Machine Intelligence}} \bibinfo{volume}{1}, \bibinfo{number}{11} (\bibinfo{date}{November} \bibinfo{year}{2019}), \bibinfo{pages}{501--507}.
\newblock
\showISSN{2522-5839}
\urldef\tempurl%
\url{https://doi.org/10.1038/s42256-019-0114-4}
\showDOI{\tempurl}


\bibitem[{NAIAC}(2023)]%
        {naiac2023aers}
\bibfield{author}{\bibinfo{person}{{NAIAC}}.} \bibinfo{year}{2023}\natexlab{}.
\newblock \bibinfo{title}{RECOMMENDATION: Improve Monitoring of Emerging Risks from AI through Adverse Event Reporting}.
\newblock
\newblock
\urldef\tempurl%
\url{https://ai.gov/wp-content/uploads/2023/12/Recommendation_Improve-Monitoring-of-Emerging-Risks-from-AI-through-Adverse-Event-Reporting.pdf}
\showURL{%
\tempurl}


\bibitem[Narayanan and Kapoor(2023)]%
        {narayanan2023transparencyreports}
\bibfield{author}{\bibinfo{person}{Arvind Narayanan} {and} \bibinfo{person}{Sayash Kapoor}.} \bibinfo{year}{2023}\natexlab{}.
\newblock \bibinfo{title}{Generative AI companies must publish transparency reports}.
\newblock
\newblock
\urldef\tempurl%
\url{https://knightcolumbia.org/blog/generative-ai-companies-must-publish-transparency-reports}
\showURL{%
\tempurl}


\bibitem[NYT(2024)]%
        {nyt2023nyt}
\bibfield{author}{\bibinfo{person}{NYT}.} \bibinfo{year}{2024}\natexlab{}.
\newblock \bibinfo{title}{THE NEW YORK TIMES COMPANY v. MICROSOFT CORPORATION, OPENAI, INC., OPENAI LP, OPENAI GP, LLC, OPENAI, LLC, OPENAI OPCO LLC, OPENAI GLOBAL LLC, OAI CORPORATION, LLC, and OPENAI HOLDINGS, LLC}.
\newblock
\newblock
\urldef\tempurl%
\url{https://nytco-assets.nytimes.com/2023/12/NYT_Complaint_Dec2023.pdf}
\showURL{%
\tempurl}


\bibitem[on~Neural Information Processing~Systems(2022)]%
        {NeurIPS2022}
\bibfield{author}{\bibinfo{person}{Conference on Neural Information Processing~Systems}.} \bibinfo{year}{2022}\natexlab{}.
\newblock \bibinfo{title}{NeurIPS 2022 Paper Checklist Guidelines}.
\newblock
\newblock
\urldef\tempurl%
\url{https://neurips.cc/Conferences/2022/PaperInformation/PaperChecklist}
\showURL{%
\tempurl}


\bibitem[on~Neural Information Processing~Systems(2023)]%
        {EMNLP2023}
\bibfield{author}{\bibinfo{person}{Conference on Neural Information Processing~Systems}.} \bibinfo{year}{2023}\natexlab{}.
\newblock \bibinfo{title}{Call for Main Conference Papers}.
\newblock
\newblock
\urldef\tempurl%
\url{https://2023.emnlp.org/calls/main_conference_papers/}
\showURL{%
\tempurl}


\bibitem[OpenAI(2023)]%
        {openai2023gpt4}
\bibfield{author}{\bibinfo{person}{OpenAI}.} \bibinfo{year}{2023}\natexlab{}.
\newblock \bibinfo{title}{GPT-4 Technical Report}.
\newblock
\newblock
\showeprint[arxiv]{2303.08774}~[cs.CL]


\bibitem[Patterson et~al\mbox{.}(2021)]%
        {patterson2021carbon}
\bibfield{author}{\bibinfo{person}{David Patterson}, \bibinfo{person}{Joseph Gonzalez}, \bibinfo{person}{Quoc Le}, \bibinfo{person}{Chen Liang}, \bibinfo{person}{Lluis-Miquel Munguia}, \bibinfo{person}{Daniel Rothchild}, \bibinfo{person}{David So}, \bibinfo{person}{Maud Texier}, {and} \bibinfo{person}{Jeff Dean}.} \bibinfo{year}{2021}\natexlab{}.
\newblock \showarticletitle{Carbon emissions and large neural network training}.
\newblock \bibinfo{journal}{\emph{arXiv preprint arXiv:2104.10350}} (\bibinfo{year}{2021}).
\newblock


\bibitem[Perino(2010)]%
        {perino2010hellhound}
\bibfield{author}{\bibinfo{person}{M. Perino}.} \bibinfo{year}{2010}\natexlab{}.
\newblock \bibinfo{booktitle}{\emph{The Hellhound of Wall Street: How Ferdinand Pecora's Investigation of the Great Crash Forever Changed American Finance}}.
\newblock \bibinfo{publisher}{Penguin Publishing Group}.
\newblock
\showISBNx{9780143120032}
\showLCCN{2010019157}
\urldef\tempurl%
\url{https://books.google.com/books?id=VJZPEAAAQBAJ}
\showURL{%
\tempurl}


\bibitem[Perrigo(2022)]%
        {perrigo2022kenya}
\bibfield{author}{\bibinfo{person}{Billy Perrigo}.} \bibinfo{year}{2022}\natexlab{}.
\newblock \showarticletitle{Exclusive: OpenAI Used Kenyan Workers on Less Than 2 Per Hour to Make ChatGPT Less Toxic}.
\newblock \bibinfo{journal}{\emph{Time}} (\bibinfo{year}{2022}).
\newblock
\urldef\tempurl%
\url{https://time.com/6247678/openai-chatgpt-kenya-workers}
\showURL{%
\tempurl}


\bibitem[Pichai and Hassabis({[n.\,d.]})]%
        {pichai2023gemini}
\bibfield{author}{\bibinfo{person}{Sundar Pichai} {and} \bibinfo{person}{Demis Hassabis}.} \bibinfo{year}{[n.\,d.]}\natexlab{}.
\newblock \bibinfo{title}{Introducing Gemini: our largest and most capable AI model}.
\newblock
\newblock


\bibitem[Raji and Buolamwini(2019)]%
        {raji2019actionable}
\bibfield{author}{\bibinfo{person}{Inioluwa~Deborah Raji} {and} \bibinfo{person}{Joy Buolamwini}.} \bibinfo{year}{2019}\natexlab{}.
\newblock \showarticletitle{Actionable Auditing: Investigating the Impact of Publicly Naming Biased Performance Results of Commercial AI Products}. In \bibinfo{booktitle}{\emph{Proceedings of the 2019 AAAI/ACM Conference on AI, Ethics, and Society}} (Honolulu, HI, USA) \emph{(\bibinfo{series}{AIES '19})}. \bibinfo{publisher}{Association for Computing Machinery}, \bibinfo{address}{New York, NY, USA}, \bibinfo{pages}{429–435}.
\newblock
\showISBNx{9781450363242}
\urldef\tempurl%
\url{https://doi.org/10.1145/3306618.3314244}
\showDOI{\tempurl}


\bibitem[Rydzak(2023)]%
        {rydzak2023stalled}
\bibfield{author}{\bibinfo{person}{Jan Rydzak}.} \bibinfo{year}{2023}\natexlab{}.
\newblock \bibinfo{title}{The Stalled Machines of Transparency Reporting}.
\newblock \bibinfo{howpublished}{\url{https://carnegieendowment.org/2023/11/29/stalled-machines-of-transparency-reporting-pub-91085}}.
\newblock


\bibitem[{Santa Clara Principles}(2023)]%
        {santaclara2023principles}
\bibfield{author}{\bibinfo{person}{{Santa Clara Principles}}.} \bibinfo{year}{2023}\natexlab{}.
\newblock \bibinfo{title}{The Santa Clara Principles: On Transparency and Accountability in Content Moderation}.
\newblock \bibinfo{howpublished}{\url{https://santaclaraprinciples.org/}}.
\newblock


\bibitem[Schatz(2006)]%
        {schatz2006tech}
\bibfield{author}{\bibinfo{person}{Amy Schatz}.} \bibinfo{year}{2006}\natexlab{}.
\newblock \bibinfo{title}{Tech Firms Defend China Web Policies}.
\newblock \bibinfo{howpublished}{\url{https://www.wsj.com/articles/SB114002162437674809}}.
\newblock


\bibitem[Schneider et~al\mbox{.}(2023)]%
        {Schneider2023Collaborative}
\bibfield{author}{\bibinfo{person}{Jens-Peter Schneider}, \bibinfo{person}{Kester Siegrist}, {and} \bibinfo{person}{Simon Oles}.} \bibinfo{year}{2023}\natexlab{}.
\newblock \showarticletitle{Collaborative Governance of the EU Digital Single Market established by the Digital Services Act}.
\newblock \bibinfo{journal}{\emph{University of Luxembourg Law Research Paper}} \bibinfo{volume}{2023}, \bibinfo{number}{09} (\bibinfo{date}{4 September} \bibinfo{year}{2023}).
\newblock
\urldef\tempurl%
\url{https://ssrn.com/abstract=4561010}
\showURL{%
\tempurl}


\bibitem[Secutiries and Commision(2024a)]%
        {SECmain}
\bibfield{author}{\bibinfo{person}{U.S. Secutiries} {and} \bibinfo{person}{Exchange Commision}.} \bibinfo{year}{2024}\natexlab{a}.
\newblock \bibinfo{title}{About the SEC}.
\newblock \bibinfo{howpublished}{\url{https://www.sec.gov/strategic-plan/about}}.
\newblock


\bibitem[Secutiries and Commision(2024b)]%
        {SECform10K}
\bibfield{author}{\bibinfo{person}{U.S. Secutiries} {and} \bibinfo{person}{Exchange Commision}.} \bibinfo{year}{2024}\natexlab{b}.
\newblock \bibinfo{title}{Form 10-K}.
\newblock \bibinfo{howpublished}{\url{https://www.investor.gov/introduction-investing/investing-basics/glossary/form-10-k}}.
\newblock


\bibitem[Secutiries and Commision(2024c)]%
        {SECform8k}
\bibfield{author}{\bibinfo{person}{U.S. Secutiries} {and} \bibinfo{person}{Exchange Commision}.} \bibinfo{year}{2024}\natexlab{c}.
\newblock \bibinfo{title}{Form 8-K}.
\newblock \bibinfo{howpublished}{\url{https://www.investor.gov/introduction-investing/investing-basics/glossary/form-8-k}}.
\newblock


\bibitem[Secutiries and Commision(2024d)]%
        {GAAP}
\bibfield{author}{\bibinfo{person}{U.S. Secutiries} {and} \bibinfo{person}{Exchange Commision}.} \bibinfo{year}{2024}\natexlab{d}.
\newblock \bibinfo{title}{Generally Accepted Accounting Principles (GAAP)}.
\newblock \bibinfo{howpublished}{\url{https://www.investor.gov/introduction-investing/investing-basics/glossary/generally-accepted-accounting-principles-gaap}}.
\newblock


\bibitem[Secutiries and Commision(2024e)]%
        {PAOCB}
\bibfield{author}{\bibinfo{person}{U.S. Secutiries} {and} \bibinfo{person}{Exchange Commision}.} \bibinfo{year}{2024}\natexlab{e}.
\newblock \bibinfo{title}{Generally Accepted Accounting Principles (GAAP)}.
\newblock \bibinfo{howpublished}{\url{https://www.investor.gov/introduction-investing/investing-basics/glossary/generally-accepted-accounting-principles-gaap}}.
\newblock


\bibitem[Stoughton and Rosenzweig(2022)]%
        {StoughtonRosenzweig2022}
\bibfield{author}{\bibinfo{person}{Katie Stoughton} {and} \bibinfo{person}{Paul Rosenzweig}.} \bibinfo{year}{2022}\natexlab{}.
\newblock \bibinfo{title}{Toward Greater Content Moderation Transparency Reporting}.
\newblock \bibinfo{howpublished}{Lawfare}.
\newblock
\urldef\tempurl%
\url{https://www.lawfaremedia.org/article/toward-greater-content-moderation-transparency-reporting}
\showURL{%
\tempurl}


\bibitem[{Trust and Safety Professional Association}(2023)]%
        {tspa2023transparency}
\bibfield{author}{\bibinfo{person}{{Trust and Safety Professional Association}}.} \bibinfo{year}{2023}\natexlab{}.
\newblock \bibinfo{title}{Transparency Reporting}.
\newblock \bibinfo{howpublished}{\url{https://www.tspa.org/curriculum/ts-fundamentals/transparency-report/}}.
\newblock


\bibitem[{UK CMA}(2023)]%
        {cma2023ai}
\bibfield{author}{\bibinfo{person}{{UK CMA}}.} \bibinfo{year}{2023}\natexlab{}.
\newblock \bibinfo{title}{AI Foundation Models: Initial Report}.
\newblock
\newblock
\urldef\tempurl%
\url{https://assets.publishing.service.gov.uk/media/65081d3aa41cc300145612c0/Full_report_.pdf}
\showURL{%
\tempurl}


\bibitem[{United States Executive Office of the President}(2023)]%
        {us2023eo}
\bibfield{author}{\bibinfo{person}{{United States Executive Office of the President}}.} \bibinfo{year}{2023}\natexlab{}.
\newblock \bibinfo{title}{Executive Order on Safe, Secure, and Trustworthy Development and Use of Artificial Intelligence}.
\newblock
\newblock
\urldef\tempurl%
\url{https://www.federalregister.gov/documents/2023/11/01/2023-24283/safe-secure-and-trustworthy-development-and-use-of-artificial-intelligence}
\showURL{%
\tempurl}


\bibitem[Urman and Makhortykh(2023)]%
        {UrmanMakhortykh2023}
\bibfield{author}{\bibinfo{person}{Aleksandra Urman} {and} \bibinfo{person}{Mykola Makhortykh}.} \bibinfo{year}{2023}\natexlab{}.
\newblock \showarticletitle{How transparent are transparency reports? Comparative analysis of transparency reporting across online platforms}.
\newblock \bibinfo{journal}{\emph{Telecommunications Policy}} \bibinfo{volume}{47}, \bibinfo{number}{3} (\bibinfo{year}{2023}), \bibinfo{pages}{102477}.
\newblock
\showISSN{0308-5961}
\urldef\tempurl%
\url{https://doi.org/10.1016/j.telpol.2022.102477}
\showDOI{\tempurl}


\bibitem[{Valerie C. Brannon and Victoria L. Killion and Whitney K. Novak and L. Paige Whitaker}(2023)]%
        {crs2023first}
\bibfield{author}{\bibinfo{person}{{Valerie C. Brannon and Victoria L. Killion and Whitney K. Novak and L. Paige Whitaker}}.} \bibinfo{year}{2023}\natexlab{}.
\newblock \bibinfo{title}{First Amendment Limitations on Disclosure Requirements}.
\newblock
\newblock
\urldef\tempurl%
\url{https://crsreports.congress.gov/product/pdf/IF/IF12388}
\showURL{%
\tempurl}


\bibitem[Vermeulen(2021)]%
        {Vermeulen2021keys}
\bibfield{author}{\bibinfo{person}{Mathias Vermeulen}.} \bibinfo{year}{2021}\natexlab{}.
\newblock \bibinfo{title}{The Keys to the Kingdom}.
\newblock
\newblock
\urldef\tempurl%
\url{https://knightcolumbia.org/content/the-keys-to-the-kingdom}
\showURL{%
\tempurl}


\bibitem[Vipra and Korinek(2023)]%
        {vipra2023concentration}
\bibfield{author}{\bibinfo{person}{Jai Vipra} {and} \bibinfo{person}{Anton Korinek}.} \bibinfo{year}{2023}\natexlab{}.
\newblock \showarticletitle{Market concentration implications of foundation models: The Invisible Hand of ChatGPT}.
\newblock \bibinfo{journal}{\emph{The Brookings Institution}} (\bibinfo{year}{2023}).
\newblock
\urldef\tempurl%
\url{https://www.brookings.edu/articles/market-concentration-implications-of-foundation-models-the-invisible-hand-of-chatgpt}
\showURL{%
\tempurl}


\bibitem[Vogus and Llansó(2021)]%
        {cdt2021}
\bibfield{author}{\bibinfo{person}{Caitlin Vogus} {and} \bibinfo{person}{Emma Llansó}.} \bibinfo{year}{2021}\natexlab{}.
\newblock \showarticletitle{Making Transparency Meaningful: A Framework for Policymakers}.
\newblock \bibinfo{journal}{\emph{Center for Democracy and Technology}} (\bibinfo{year}{2021}).
\newblock
\urldef\tempurl%
\url{https://cdt.org/insights/report-making-transparency-meaningful-a-framework-for-policymakers/}
\showURL{%
\tempurl}


\bibitem[Wei et~al\mbox{.}(2022)]%
        {wei2022emergent}
\bibfield{author}{\bibinfo{person}{Jason Wei}, \bibinfo{person}{Yi Tay}, \bibinfo{person}{Rishi Bommasani}, \bibinfo{person}{Colin Raffel}, \bibinfo{person}{Barret Zoph}, \bibinfo{person}{Sebastian Borgeaud}, \bibinfo{person}{Dani Yogatama}, \bibinfo{person}{Maarten Bosma}, \bibinfo{person}{Denny Zhou}, \bibinfo{person}{Donald Metzler}, \bibinfo{person}{Ed~H. Chi}, \bibinfo{person}{Tatsunori Hashimoto}, \bibinfo{person}{Oriol Vinyals}, \bibinfo{person}{Percy Liang}, \bibinfo{person}{Jeff Dean}, {and} \bibinfo{person}{William Fedus}.} \bibinfo{year}{2022}\natexlab{}.
\newblock \showarticletitle{Emergent Abilities of Large Language Models}.
\newblock \bibinfo{journal}{\emph{Transactions on Machine Learning Research}} (\bibinfo{year}{2022}).
\newblock
\urldef\tempurl%
\url{https://openreview.net/forum?id=yzkSU5zdwD}
\showURL{%
\tempurl}
\newblock
\shownote{Survey Certification}.


\bibitem[Weidinger et~al\mbox{.}(2022)]%
        {weidinger2022taxonomy}
\bibfield{author}{\bibinfo{person}{Laura Weidinger}, \bibinfo{person}{Jonathan Uesato}, \bibinfo{person}{Maribeth Rauh}, \bibinfo{person}{Conor Griffin}, \bibinfo{person}{Po-Sen Huang}, \bibinfo{person}{John Mellor}, \bibinfo{person}{Amelia Glaese}, \bibinfo{person}{Myra Cheng}, \bibinfo{person}{Borja Balle}, \bibinfo{person}{Atoosa Kasirzadeh}, \bibinfo{person}{Courtney Biles}, \bibinfo{person}{Sasha Brown}, \bibinfo{person}{Zac Kenton}, \bibinfo{person}{Will Hawkins}, \bibinfo{person}{Tom Stepleton}, \bibinfo{person}{Abeba Birhane}, \bibinfo{person}{Lisa~Anne Hendricks}, \bibinfo{person}{Laura Rimell}, \bibinfo{person}{William Isaac}, \bibinfo{person}{Julia Haas}, \bibinfo{person}{Sean Legassick}, \bibinfo{person}{Geoffrey Irving}, {and} \bibinfo{person}{Iason Gabriel}.} \bibinfo{year}{2022}\natexlab{}.
\newblock \showarticletitle{Taxonomy of Risks Posed by Language Models}. In \bibinfo{booktitle}{\emph{2022 ACM Conference on Fairness, Accountability, and Transparency}} (Seoul, Republic of Korea) \emph{(\bibinfo{series}{FAccT '22})}. \bibinfo{publisher}{Association for Computing Machinery}, \bibinfo{address}{New York, NY, USA}, \bibinfo{pages}{214–229}.
\newblock
\showISBNx{9781450393522}
\urldef\tempurl%
\url{https://doi.org/10.1145/3531146.3533088}
\showDOI{\tempurl}


\bibitem[Williams et~al\mbox{.}(2022)]%
        {williams2022exploited}
\bibfield{author}{\bibinfo{person}{Adrienne Williams}, \bibinfo{person}{Milagros Miceli}, {and} \bibinfo{person}{Timnit Gebru}.} \bibinfo{year}{2022}\natexlab{}.
\newblock \bibinfo{booktitle}{\emph{De Anima: On the Soul}}.
\newblock
\urldef\tempurl%
\url{https://www.noemamag.com/the-exploited-labor-behind-artificial-intelligence/}
\showURL{%
\tempurl}


\bibitem[{X}(2023)]%
        {twitter2023update}
\bibfield{author}{\bibinfo{person}{{X}}.} \bibinfo{year}{2023}\natexlab{}.
\newblock \bibinfo{title}{An update on Twitter Transparency Reporting}.
\newblock \bibinfo{howpublished}{\url{https://blog.twitter.com/en_us/topics/company/2023/an-update-on-twitter-transparency-reporting}}.
\newblock


\bibitem[Zalnieriute(2021)]%
        {zalnieriute2021transparency}
\bibfield{author}{\bibinfo{person}{Monika Zalnieriute}.} \bibinfo{year}{2021}\natexlab{}.
\newblock \bibinfo{booktitle}{\emph{\textquotedblleft{}Transparency-Washing\textquotedblright{} in the Digital Age : A Corporate Agenda of Procedural Fetishism}}.
\newblock \bibinfo{type}{{T}echnical {R}eport}.
\newblock
\urldef\tempurl%
\url{http://hdl.handle.net/11159/468588}
\showURL{%
\tempurl}


\bibitem[Zhu(2015)]%
        {zhu2015perfect}
\bibfield{author}{\bibinfo{person}{Jenny Zhu}.} \bibinfo{year}{2015}\natexlab{}.
\newblock \bibinfo{title}{A perfect EFF score! We're proud to have your back}.
\newblock \bibinfo{howpublished}{\url{https://wordpress.com/blog/2015/06/17/a-perfect-eff-score-were-proud-to-have-your-back/}}.
\newblock


\end{thebibliography}
\clearpage
\appendix
\hypertarget{indicators}{\section{Indicators}}
\label{app:indicators}

\begin{figure}
\centering
\includegraphics[keepaspectratio, height=0.9\textheight, width=0.9\textwidth]{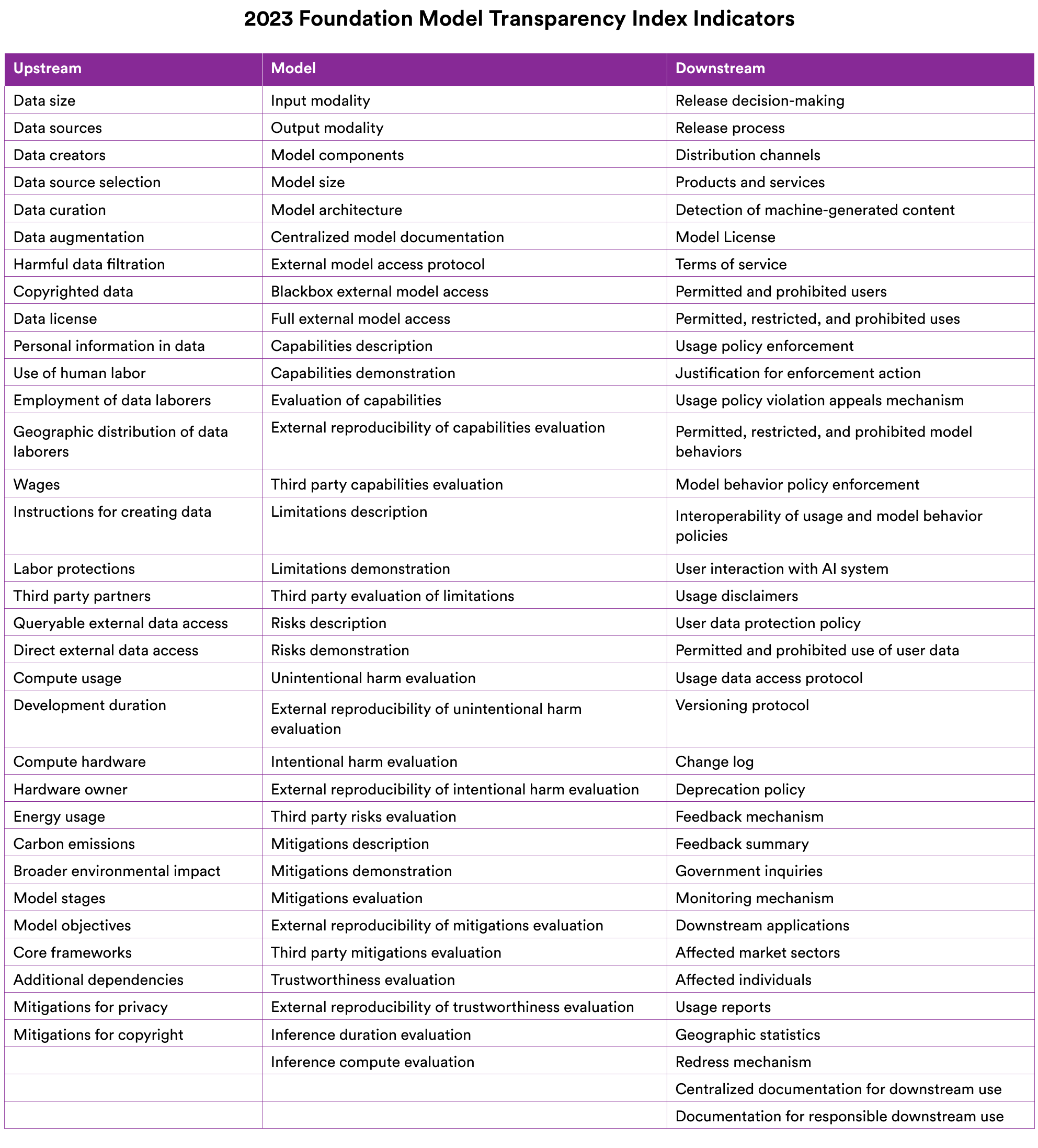}
\caption{\textbf{Indicators.} The \numindicators indicators from the \projectname spanning the \numdomains domains: upstream, model, and downstream; figure used with permission from \citet{bommasani2023foundation}.
% Would be good to remove horizontal lines
}.
\label{fig:indicators}
\end{figure}

We use the 100 transparency indicators from the Foundation Model Transparency Index \citep{bommasani2023foundation} as the basis for our transparency reports. 
These indicators are listed by name in \autoref{fig:indicators} with definitions available at \url{https://github.com/stanford-crfm/fmti/blob/main/fmti-indicators.csv}. 

\clearpage
% Note: Update once matrix is added
\hypertarget{alignment}{\section{Policy Alignment}}
\label{app:alignment}

We track transparency requirements for foundation model developers in 6 government policies from Canada, the EU, the US, and the G7.
For each of our 100 transparency indicators, we indicate if the associated policy has a transparency requirement that addresses the same matter as our indicator.
We consider a transparency requirement to be aligned with one of our transparency indicators if (i) the requirement addresses the same issue area as the indicator, (ii) the requirement is directed to developers of foundation models, and (iii) the policy explicitly requires transparency or information sharing in this area.\footnote{We were relatively conservative in stating which requirements align with our transparency indicators, erring on the side of caution in regulatory interpretation as there is little to no available evidence regarding how these recent policies will be interpreted or enforced in practice.}
We used a single annotator for each policy, who was responsible for comparing each transparency requirement in the policy to each transparency indicator in the Foundation Model Transparency Index.
There are several important caveats regarding the extent of this overlap.
Whereas foundation model transparency reports are intended to be documents that are publicly available, transparency requirements in government policies may only require foundation model developers to disclose information to the government or to other firms (e.g. downstream developers). 
Moreover, the voluntary government policies we consider impose high-level transparency requirements, meaning it is difficult to discern whether they correspond precisely to a narrow indicator of transparency.
The resulting $100 \times 6$ matrix of (indicator, policy) pairs is presented in \autoref{tab:alignment-matrix}.

\begin{longtblr}[
  caption = {\textbf{Policy Alignment Matrix for Transparency Indicators}},
  label = {tab:alignment-matrix},
]{
  colspec = {|Xcccccc|},
  rowhead = 1,
  hlines,
  row{even} = {gray9},
  row{1} = {olive9},
} 
    
        & \textbf{US WHVC} & \textbf{US EO} & \textbf{US FMTA} & \textbf{EU AIA} & \textbf{G7 CoC} & \textbf{CA CoC} \\ 
        \hline        
\textbf{Data size} & \xmark & \xmark & \cmark & \cmark & \xmark & \xmark \\ 
        \textbf{Data sources} & \xmark & \xmark & \cmark & \cmark & \xmark & \cmark \\ 
        \textbf{Data creators} & \xmark & \xmark & \cmark & \xmark & \xmark & \xmark \\ 
        \textbf{Data source selection} & \xmark & \xmark & \xmark & \cmark & \xmark & \xmark \\ 
        \textbf{Data curation} & \xmark & \xmark & \cmark & \cmark & \xmark & \xmark \\ 
        \textbf{Data augmentation} & \xmark & \xmark & \cmark & \xmark & \xmark & \xmark \\ 
        \textbf{Harmful data filtration} & \xmark & \xmark & \cmark & \cmark & \xmark & \xmark \\ 
        \textbf{Copyrighted data} & \xmark & \xmark & \cmark & \xmark & \xmark & \xmark \\ 
        \textbf{Data license} & \xmark & \xmark & \cmark & \xmark & \xmark & \xmark \\ 
        \textbf{Personal information in data} & \xmark & \xmark & \cmark & \xmark & \xmark & \xmark \\ 
        \textbf{Use of human labor} & \xmark & \xmark & \xmark & \xmark & \xmark & \xmark \\ 
        \textbf{Employment of data laborers} & \xmark & \xmark & \xmark & \xmark & \xmark & \xmark \\ 
        \textbf{Geographic distribution of data laborers} & \xmark & \xmark & \xmark & \xmark & \xmark & \xmark \\ 
        \textbf{Wages} & \xmark & \xmark & \xmark & \xmark & \xmark & \xmark \\ 
        \textbf{Instructions for creating data} & \xmark & \xmark & \xmark & \xmark & \xmark & \xmark \\ 
        \textbf{Labor protections} & \xmark & \xmark & \xmark & \xmark & \xmark & \xmark \\ 
        \textbf{Third party partners} & \xmark & \xmark & \xmark & \xmark & \xmark & \xmark \\ 
        \textbf{Queryable external data access} & \xmark & \xmark & \xmark & \xmark & \xmark & \xmark \\ 
        \textbf{Direct external data access} & \xmark & \xmark & \xmark & \xmark & \xmark & \xmark \\ 
        \textbf{Compute usage} & \xmark & \xmark & \cmark & \cmark & \xmark & \xmark \\ 
        \textbf{Development duration} & \xmark & \xmark & \xmark & \cmark & \xmark & \xmark \\ 
        \textbf{Compute hardware} & \xmark & \xmark & \xmark & \xmark & \xmark & \xmark \\ 
        \textbf{Hardware owner} & \xmark & \xmark & \xmark & \xmark & \xmark & \xmark \\ 
        \textbf{Energy usage} & \xmark & \xmark & \xmark & \cmark & \xmark & \xmark \\ 
        \textbf{Carbon emissions} & \xmark & \xmark & \xmark & \xmark & \xmark & \xmark \\ 
        \textbf{Broader environmental impact} & \xmark & \xmark & \xmark & \xmark & \xmark & \xmark \\ 
        \textbf{Model stages} & \xmark & \xmark & \xmark & \xmark & \xmark & \xmark \\ 
        \textbf{Model objectives} & \xmark & \xmark & \xmark & \cmark & \xmark & \xmark \\ 
        \textbf{Core frameworks} & \xmark & \xmark & \xmark & \xmark & \xmark & \xmark \\ 
        \textbf{Additional dependencies} & \xmark & \xmark & \xmark & \xmark & \xmark & \xmark \\ 
        \textbf{Mitigations for privacy} & \xmark & \xmark & \xmark & \xmark & \xmark & \xmark \\ 
        \textbf{Mitigations for copyright} & \xmark & \xmark & \xmark & \xmark & \xmark & \xmark \\ 
        \textbf{Input modality} & \xmark & \xmark & \xmark & \cmark & \xmark & \xmark \\ 
        \textbf{Output modality} & \xmark & \xmark & \xmark & \cmark & \xmark & \xmark \\ 
        \textbf{Model components} & \xmark & \xmark & \xmark & \cmark & \xmark & \xmark \\ 
        \textbf{Model size} & \xmark & \xmark & \xmark & \cmark & \xmark & \xmark \\ 
        \textbf{Model architecture} & \xmark & \xmark & \xmark & \cmark & \xmark & \xmark \\ 
        \textbf{Centralized model documentation} & \cmark & \xmark & \xmark & \cmark & \cmark & \xmark \\ 
        \textbf{External model access protocol} & \xmark & \cmark & \xmark & \xmark & \xmark & \xmark \\ 
        \textbf{Blackbox external model access} & \xmark & \xmark & \xmark & \xmark & \xmark & \xmark \\ 
        \textbf{Full external model access} & \xmark & \xmark & \xmark & \xmark & \xmark & \xmark \\ 
        \textbf{Capabilities description} & \cmark & \xmark & \xmark & \cmark & \cmark & \cmark \\ 
        \textbf{Capabilities demonstration} & \xmark & \xmark & \xmark & \xmark & \xmark & \xmark \\ 
        \textbf{Evaluation of capabilities} & \xmark & \xmark & \xmark & \cmark & \xmark & \xmark \\ 
        \textbf{External reproducibility of capabilities evaluation} & \xmark & \xmark & \xmark & \xmark & \xmark & \xmark \\ 
        \textbf{Third party capabilities evaluation} & \xmark & \xmark & \xmark & \xmark & \xmark & \xmark \\ 
        \textbf{Limitations description} & \cmark & \xmark & \cmark & \cmark & \cmark & \cmark \\ 
        \textbf{Limitations demonstration} & \xmark & \xmark & \xmark & \xmark & \xmark & \xmark \\ 
        \textbf{Third party evaluation of limitations} & \xmark & \xmark & \xmark & \xmark & \xmark & \xmark \\ 
        \textbf{Risks description} & \cmark & \xmark & \cmark & \cmark & \cmark & \xmark \\ 
        \textbf{Risks demonstration} & \xmark & \xmark & \xmark & \xmark & \xmark & \xmark \\ 
        \textbf{Unintentional harm evaluation} & \cmark & \cmark & \cmark & \xmark & \cmark & \xmark \\ 
        \textbf{External reproducibility of unintentional harm evaluation} & \xmark & \xmark & \xmark & \xmark & \xmark & \xmark \\ 
        \textbf{Intentional harm evaluation} & \cmark & \cmark & \cmark & \cmark & \cmark & \xmark \\ 
        \textbf{External reproducibility of intentional harm evaluation} & \xmark & \xmark & \xmark & \xmark & \xmark & \xmark \\ 
        \textbf{Third party risks evaluation} & \xmark & \xmark & \cmark & \cmark & \xmark & \xmark \\ 
        \textbf{Mitigations description} & \xmark & \cmark & \cmark & \cmark & \cmark & \cmark \\ 
        \textbf{Mitigations demonstration} & \xmark & \xmark & \xmark & \xmark & \xmark & \xmark \\ 
        \textbf{Mitigations evaluation} & \xmark & \xmark & \xmark & \xmark & \xmark & \xmark \\ 
        \textbf{External reproducibility of mitigations evaluation} & \xmark & \xmark & \xmark & \xmark & \xmark & \xmark \\ 
        \textbf{Third party mitigations evaluation} & \xmark & \xmark & \xmark & \xmark & \xmark & \xmark \\ 
        \textbf{Trustworthiness evaluation} & \xmark & \xmark & \xmark & \xmark & \xmark & \xmark \\ 
        \textbf{External reproducibility of trustworthiness evaluation} & \xmark & \xmark & \xmark & \xmark & \xmark & \xmark \\ 
        \textbf{Inference duration evaluation} & \xmark & \xmark & \xmark & \xmark & \xmark & \xmark \\ 
        \textbf{Inference compute evaluation} & \xmark & \xmark & \cmark & \xmark & \xmark & \xmark \\ 
        \textbf{Release decision-making} & \xmark & \xmark & \xmark & \xmark & \xmark & \xmark \\ 
        \textbf{Release process} & \xmark & \xmark & \xmark & \xmark & \xmark & \xmark \\ 
        \textbf{Distribution channels} & \xmark & \cmark & \xmark & \cmark & \xmark & \xmark \\ 
        \textbf{Products and services} & \xmark & \xmark & \xmark & \xmark & \xmark & \xmark \\ 
        \textbf{Detection of machine-generated content} & \xmark & \xmark & \xmark & \cmark & \xmark & \cmark \\ 
        \textbf{Model License} & \xmark & \xmark & \xmark & \cmark & \xmark & \xmark \\ 
        \textbf{Terms of service} & \xmark & \xmark & \xmark & \xmark & \xmark & \xmark \\ 
        \textbf{Permitted and prohibited users} & \xmark & \xmark & \xmark & \xmark & \xmark & \xmark \\ 
        \textbf{Permitted, restricted, and prohibited uses} & \cmark & \xmark & \xmark & \cmark & \cmark & \xmark \\ 
        \textbf{Usage policy enforcement} & \xmark & \xmark & \xmark & \xmark & \xmark & \xmark \\ 
        \textbf{Justification for enforcement action} & \xmark & \xmark & \xmark & \xmark & \xmark & \xmark \\ 
        \textbf{Usage policy violation appeals mechanism} & \xmark & \xmark & \xmark & \xmark & \xmark & \xmark \\ 
        \textbf{Permitted, restricted, and prohibited model behaviors} & \xmark & \xmark & \xmark & \xmark & \xmark & \xmark \\ 
        \textbf{Model behavior policy enforcement} & \xmark & \xmark & \xmark & \xmark & \xmark & \xmark \\ 
        \textbf{Interoperability of usage and model behavior policies} & \xmark & \xmark & \xmark & \xmark & \xmark & \xmark \\ 
        \textbf{User interaction with AI system} & \xmark & \xmark & \xmark & \cmark & \cmark & \cmark \\ 
        \textbf{Usage disclaimers} & \xmark & \xmark & \xmark & \xmark & \xmark & \xmark \\ 
        \textbf{User data protection policy} & \xmark & \xmark & \cmark & \xmark & \cmark & \xmark \\ 
        \textbf{Permitted and prohibited use of user data} & \xmark & \xmark & \xmark & \xmark & \xmark & \xmark \\ 
        \textbf{Usage data access protocol} & \xmark & \xmark & \xmark & \xmark & \xmark & \xmark \\ 
        \textbf{Versioning protocol} & \xmark & \xmark & \cmark & \xmark & \xmark & \xmark \\ 
        \textbf{Change log} & \xmark & \xmark & \cmark & \xmark & \xmark & \xmark \\ 
        \textbf{Deprecation policy} & \xmark & \xmark & \xmark & \xmark & \xmark & \xmark \\ 
        \textbf{Feedback mechanism} & \xmark & \xmark & \xmark & \xmark & \xmark & \cmark \\ 
        \textbf{Feedback summary} & \xmark & \xmark & \xmark & \xmark & \xmark & \xmark \\ 
        \textbf{Government inquiries} & \xmark & \xmark & \xmark & \xmark & \xmark & \xmark \\ 
        \textbf{Monitoring mechanism} & \xmark & \xmark & \xmark & \xmark & \xmark & \cmark \\ 
        \textbf{Downstream applications} & \xmark & \xmark & \xmark & \xmark & \xmark & \xmark \\ 
        \textbf{Affected market sectors} & \xmark & \xmark & \xmark & \xmark & \xmark & \xmark \\ 
        \textbf{Affected individuals} & \xmark & \xmark & \xmark & \xmark & \xmark & \xmark \\ 
        \textbf{Usage reports} & \xmark & \xmark & \xmark & \cmark & \xmark & \xmark \\ 
        \textbf{Geographic statistics} & \xmark & \xmark & \xmark & \xmark & \xmark & \xmark \\ 
        \textbf{Redress mechanism} & \xmark & \xmark & \xmark & \xmark & \xmark & \xmark \\ 
        \textbf{Centralized documentation for downstream use} & \xmark & \xmark & \xmark & \cmark & \cmark & \xmark \\ 
        \textbf{Documentation for responsible downstream use} & \xmark & \xmark & \xmark & \cmark & \cmark & \cmark \\ 
        \textbf{Totals} & 7 & 5 & 20 & 30 & 12 & 9 \\ 
\end{longtblr}

\clearpage
\hypertarget{example}{\section{Example of Transparency Report Entries}}
\label{app:example}

To demonstrate how a transparency report may be prepared, we provide the following examples of transparency report entries.
Given the poor transparency documented in the foundation model ecosystem at present \citep{bommasani2023foundation}, we construct this document by stitching together practices across several major foundation model developers.
In doing so, our objective is to highlight a larger range of practices to give greater guidance on the basis of this example.
Further, since we constructed these entries given public information as in \citep{bommasani2023foundation}, we specifically highlight that the extent to which the information is contextualized and the methodology is clear can be significantly improved.
(For some indicators, we defer to other materials because the associated information is quite lengthy/cumbersome to provide here.)

\begin{longtblr}[
  caption = {\textbf{Example of a Foundation Model Transparency Report}},
  label = {tab:example},
]{
  colspec = {|XXX|},
  rowhead = 1,
  hlines,
  row{even} = {gray9},
  row{1} = {olive9},
} 
Indicator & Developer, Model & Value \\
Data size & Hugging Face/BigScience, BLOOMZ & 363B tokens \\
Data sources & Hugging Face/BigScience, BLOOMZ & ROOTS and xP3 \\
Data source selection & Hugging Face/BigScience, BLOOMZ & See the ROOTS paper for details on source selection and the BLOOMZ paper for details on source selection for xP3 \\
Data curation & Hugging Face/BigScience, BLOOMZ & See the ROOTS paper for details on data curation and the BLOOMZ paper for details on data curation for xP3 \\
Data augmentation & Hugging Face/BigScience, BLOOMZ & See the BLOOMZ paper for details on how P3 was augmented to produce xP3 \\
Harmful data filtration & Hugging Face/BigScience, BLOOMZ & Illegal content is filtered from LAION-5B using a CLIP-based filter; offensive examples are tagged rather than filtered using QF16 and a new sexualized content classifier, both derived from CLIP embeddings; the subset of LAION-5B that is used is further filtered using LAION’s NSFW detector with $p_{unsafe}$ = 0.1 \\
Use of human labor & Meta, Llama 2 & See the Llama 2 technical report for details on the use of human labor for fine-tuning, red-teaming, and safety evaluations (e.g. pp. 28) \\
Wages & Anthropic, Claude 2 & MTurkers were paid by task and are given “frequent bonuses.” Upworker annotators “were paid significantly above the minimum wage in California” \\
Instructions for creating data & Anthropic, Claude 2 & Instructions provided on pp. 64-66 of the Training a Helpful and Harmless Assistant with RLHF paper \\
Labor protections & OpenAI, GPT-4 & "With all workers, we follow industry-best practices by ensuring every annotator retains the right to opt out of any task they find unpleasant, receive a market wage commensurate with the work they deliver, and have opportunities and channels through which they can discuss their work and raise objections. ... For sensitive content annotation, we use vendor-provided features like mandated breaks, blurring or grayscale of materials, and clearly delineated project categories such that no contractor is surprised by the nature of the material. Additionally, for vendor-managed workers, we have implemented ongoing workers’ wellness surveys and support procedures that we regularly discuss with our vendor" \\
Third party partners & Hugging Face/BigScience, BLOOMZ & BigScience data catalogue includes details regarding contributors to data crowdsourcing efforts \\
Queryable external data access & Hugging Face/BigScience, BLOOMZ & ROOTS is queryable via a tool built for precisely this type of access and xP3 is released publicly \\
Direct external data access & Hugging Face/BigScience, BLOOMZ & Full access to ROOTS is available via a form and xP3 is released publicly \\
Development duration & Meta, Llama 2 & 3.3M hours \\
Compute hardware & Stability AI, Stable Diffusion 2 & 256 NVIDIA A100 40GB GPUs \\
Hardware owner & Stability AI, Stable Diffusion 2 & Amazon Web Services \\
Energy usage & Meta, Llama 2 & 1 * $10^9$ mWh \\
Carbon emissions & Meta, Llama 2 & 539tC02 \\
Model stages & Stability AI, Stable Diffusion 2 & Training procedure described in detail in the model card.  \\
Model objectives & Meta, Llama 2 & Next word prediction as described in LLaMA 1 for pretraining, an autoregressive objective only for answer tokens for fine-tuning, reward model for RLHF. \\
Core frameworks & Hugging Face/BigScience, BLOOMZ & Frameworks for BLOOM include Megatron-DeepSpeed for large-scale distributed training, PyTorch for overall deep learning framework, and apex for FP16 and further frameworks for BLOOMZ adaptation are made clear via code release \\
Additional dependencies & Google, PaLM 2 & No additional dependencies \\
Mitigations for privacy & Meta, Llama 2 & We excluded data from certain sites known to contain a high volume of personal information about private individuals \\
Mitigations for copyright & Hugging Face/BigScience, BLOOMZ & Crowdsourcers for data that was included in ROOTS were encouraged to make "an effort to collect sources with an open license or without copyright" \\
Input modality & AI21 Labs, Jurassic-2 & Text \\
Output modality & AI21 Labs, Jurassic-2 & Text \\
Model components & Meta, Llama 2 & The model is a single component based on the Transformer architecture \\
Model size & Hugging Face/BigScience, BLOOMZ & The model size is 176B parameters (dense model) with 3.6B embedding parameters \\
Model architecture & Hugging Face/BigScience, BLOOMZ & BLOOMZ employs a transformer decoder-only model architecture \\
Centralized model documentation & Google, PaLM 2 & The technical report and model card therein provides centralized documentation \\
External model access protocol & OpenAI, GPT-4 & Research access program allows external entities to request access---a decision of will be granted within 4-6 weeks \\
Blackbox external model access & OpenAI, GPT-4 & OpenAI API provides blackbox access \\
Full external model access & Meta, Llama 2 & Model weights are openly accessible to those who agree to the custom commercial license \\
Capabilities description & Stability AI, Stable Diffusion 2 & Generating novel images from text or a preexisting image, enhancing the resolution of images, and inpainting \\
Capabilities demonstration & OpenAI, GPT-4 & Demonstrations of question answering, text completion, and various other capabilities included in the technical report \\
Evaluation of capabilities & Meta, Llama 2 & Evauations on benchmarks including MMLU, HellaSwag, and Human-Eval provided in the technical report \\
External reproducibility of capabilities evaluation & Meta, Llama 2 & Hyperparameters and prompting information are provided for capabilities evaluations, enhancing reproducibility on standard benchmarks \\
Third party capabilities evaluation & Cohere, Command & Stanford’s Center for Research on Foundation Models conducted a third-party evaluation of capabilities on the HELM benchmark \\
Limitations description & Inflection, Inflection-1 & Known limitations include hallucinations, bias, and limited non-English support \\
Limitations demonstration & Cohere, Command & Examples are provided for limitations related to non-English language support, bias, and factuality \\
Third party evaluation of limitations & Hugging Face/BigScience, BLOOMZ & The weights of the model are openly released, meaning third parties can evaluate limitations. \\
Risks description & OpenAI, GPT-4 & Risks include generating harmful content such as hate speech, amplifying bias, and generating vulnerable code \\
Risks demonstration & OpenAI, GPT-4 & System card demonstrates various types of risks \\
Unintentional harm evaluation & Google, PaLM 2 & See figure 31 in the technical report for evaluations of the rate of toxic continuation across languages or Table 24 for translation misgendering in Table 24 \\
External reproducibility of unintentional harm evaluation & Google, PaLM 2 & Toxicity evaluations and translation evaluations are reproducible, involving public datasets like ParlAI with stated hyperparameters for sampling and the Perspective API for toxicity classification with a fixed version of the Perspective API. \\
External reproducibility of intentional harm evaluation & OpenAI, GPT-4 & The evaluations in Appendix E are reproducible (prompts are provided in Figure 10). \\
Third party risks evaluation & Cohere, Command & Stanford’s Center for Research on Foundation Models conducted a third-party evaluation of risks on the HELM benchmark \\
Mitigations description & OpenAI, GPT-4 & See section 6 of the technical report for details on adversarial testing via domain experts and the model-assisted safety pipeline \\
Mitigations demonstration & OpenAI, GPT-4 & See section 2 of the system card for examples of various mitigations \\
Mitigations evaluation & OpenAI, GPT-4 & See section 3 of the system card for quantitative evaluations of the RLHF and rule-based reward models \\
Trustworthiness evaluation & OpenAI, GPT-4 & See section 6 of the technical report for a calibration evaluation \\
Inference duration evaluation & Meta, Llama 2 & Inference duration evaluated in figure 24 of technical report \\
Release decision-making & Hugging Face/BigScience, BLOOM & See section 3.6 of the BLOOM paper for discussion of release decision-making---we chose to strike a balance between unrestricted open-access and responsible-use \\
Release process & Stability AI, Stable Diffusion 2 & Model weights are made fully available upon release with no intermediate steps. \\
Distribution channels & Stability AI, Stable Diffusion 2 & Distribution channels include downloading model weights on Hugging Face and accessing the model via the Stability API \\
Products and services & Anthropic, Claude 2 & Claude and Claude Instant are powered by Claude 2 \\
Detection of machine-generated content & Stability AI, Stable Diffusion 2 & Stable Diffusion 2 incorporates an invisible watermark to help viewers identify images as machine generated \\
Model License & Hugging Face/BigScience, BLOOMZ & BigScience RAIL License \\
Terms of service & Inflection, Inflection-1 & Terms of Service for Inflection-1 and other services are available here \\
Permitted and prohibited users & Google, PaLM 2 & PaLM API additional Terms of Service include details on geographies where it is available \\
Permitted, restricted, and prohibited uses & Anthropic, Claude 2 & Acceptable Use Policy details prohibited uses such as generating content that is abusive or fraudulent content, sexual, or violent \\
Usage policy enforcement & Anthropic, Claude 2 & If we discover that your product or usage violates Anthropic’s policies, we may issue a warning requesting a change in your behavior, adjust the safety settings of your in-product experience, or suspend your access to our tools and services. \\
Justification for enforcement action & Google, PaLM 2 & If content was blocked, the response from the API contains the reason it was blocked in the ContentFilter.reason field. \\
Usage policy violation appeals mechanism & OpenAI, GPT-4 & Violations of the usage policy may present users with the opportunity to appeal via the following form \\
Permitted, restricted, and prohibited model behaviors & Anthropic, Claude 2 & The constitution we give our AI systems describes permitted and prohibited model behaviors \\
Model behavior policy enforcement & Inflection, Inflection-1 & See the review and improvement portion of the safety policy \\
Interoperability of usage and model behavior policies & OpenAI, GPT-4 & We reduced the prevalence of certain kinds of content that violate our usage policies (such as inappropriate erotic content) in our pre-training dataset, and fine-tuned the model to refuse certain instructions such as direct requests for illicit advice. \\
User interaction with AI system & OpenAI, GPT-4 & ChatGPT Plus references at the top of the page that the user is interacting with GPT-4 and at the bottom of the page links to the release notes with details about the specific version. \\
Usage disclaimers & Anthropic, Claude 2 & Users are provided with the acceptable usage policy upon making an account  \\
User data protection policy & Cohere, Command & See the privacy policy and the data usage policy \\
Permitted and prohibited use of user data & Cohere, Command & See the privacy policy and the data usage policy \\
Versioning protocol & OpenAI, GPT-4 & Each model version is dated with an -MMDD suffix; e.g., gpt-4-0613 \\
Change log & Stability AI, Stable Diffusion 2 & See the news subsection on GitHub \\
Deprecation policy & Google, PaLM 2 & Each stable version of the model is available for 6 months after the release of the next stable version \\
Feedback mechanism & Meta, Llama 2 & Report risky content generated by the model using the Llama output feedback form \\
Monitoring mechanism & Inflection, Inflection-1 & We automate monitoring across our platform to understand usage, conversational quality, and where our models might be failing to meet our safety policy. \\
Centralized documentation for downstream use & AI21 Labs, Jurassic-2 & See API reference \\
Documentation for responsible downstream use & Meta, Llama 2 & See Llama 2 responsible use guide 
\end{longtblr}

\end{document}